\documentclass[fleqn,10pt]{wlscirep}
\usepackage[utf8]{inputenc}
\usepackage[T1]{fontenc}
\usepackage{multirow}
\usepackage{multicol}
\usepackage{adjustbox}
\usepackage{etoolbox}
\pretocmd{\abstract}{\newpage}{}{}
\usepackage{newfloat}
\DeclareFloatingEnvironment[name={Supplementary Figure},fileext=lof]{suppfigure}
\DeclareFloatingEnvironment[name={Supplementary Table},fileext=lof]{suppTable}
\usepackage{lscape}
\usepackage{fltpage}
\usepackage{xcolor} 
\usepackage{soul,color}
\usepackage{rotating}
\title{Assessing generalisability of deep learning-based polyp detection and segmentation methods through a computer vision challenge}
\author[1,2,3,*]{Sharib Ali}
\author[5]{Noha Ghatwary}
\author[6,7]{Debesh Jha}
\author[13]{Ece Isik-Polat}
\author[13]{Gorkem Polat}
\author[14]{Chen YANG}
\author[14]{Wuyang LI}
\author[15]{Adrian Galdran}
\author[15]{Miguel-Ángel González Ballester}
\author[6]{Vajira Thambawita}
\author[6]{Steven Hicks}
\author[16]{Sahadev Poudel}
\author[16]{Sang-Woong Lee}
\author[17]{Ziyi Jin}
\author[17]{Tianyuan Gan}
\author[18]{ChengHui Yu}  
\author[19]{JiangPeng Yan}
\author[20]{Doyeob Yeo}
\author[21]{Hyunseok Lee}
 \author[22]{Nikhil Kumar Tomar}
\author[23]{Mahmood Haithmi}
\author[23]{Amr Ahmed}

\author[6,7]{Michael A. Riegler}
\author[24]{Christian Daul}
\author[6,25]{{P{\aa}l} Halvorsen}
\author[1,3]{Jens Rittscher}
\author[12]{Osama E. Salem}
\author[11]{Dominique {Lamarque}}
\author[10,$\dag$]{Renato {Cannizzaro}}
\author[9,$\dag$]{Stefano {Realdon}}
\author[8,26,27,$\dag$]{Thomas de Lange}
\author[2,4,$\dag$]{James E. East}
\affil[1]{Institute of Biomedical Engineering, Department of Engineering Science, University of Oxford, OX3 7DQ, Oxford, United Kingdom}
\affil[2]{Oxford National Institute for Health Research Biomedical Research Centre, OX4 2PG, Oxford, United Kingdom}
\affil[3]{Big Data Institute, University of Oxford, Li Ka Shing Centre for Health Information and Discovery, OX3 7LF, Oxford, United Kingdom}
\affil[4]{Translational Gastroenterology Unit, Nuffield Department of Medicine, Experimental Medicine Div., John Radcliffe Hospital, University of Oxford, OX3 9DU, Oxford, United Kingdom}
\affil[5]{Computer Engineering Department, Arab Academy for Science and Technology, 1029, Alexandria, Egypt}
\affil[6]{SimulaMet, 0167 Oslo, Norway}
\affil[7]{Department of Computer Science, UiT The Arctic University of Norway, Hansine Hansens veg 18, 9019 Tromsø, Norway}
\affil[8]{Medical Department, Sahlgrenska University Hospital-Mölndal, Blå stråket 5, 413 45 Göteborg, Sweden}
\affil[9]{Veneto Institute of Oncology IOV-IRCCS, Via Gattamelata, 64, 35128 Padua, Italy}
\affil[10]{CRO Centro Riferimento Oncologico IRCCS Aviano Italy, Via Franco Gallini, 2, 33081 Aviano PN, Italy}
\affil[11]{Universit{\'e} de Versailles St-Quentin en Yvelines, H{\^o}pital Ambroise Par{\'e}, 9 Av. Charles de Gaulle, 92100 Boulogne-Billancourt, France}
\affil[12]{Faculty of Medicine, University of Alexandria, 21131, Alexandria, Egypt}
\affil[13]{Graduate School of Informatics, Middle East Technical University, 06800 Ankara, Turkey}
\affil[14]{City University of Hong Kong,Kowloon, Hong Kong}
\affil[15]{BCN MedTech, Dept. of Information and Communication Technologies, Universitat Pompeu Fabra, 08018, Barcelona, Spain}
\affil[16]{Department of IT Convergence Engineering, Gachon University, Seongnam 13120, Republic of Korea}
\affil[17]{College of Biomedical Engineering and Instrument Science, Zhejiang University, Hangzhou 310027, China}
\affil[18]{Tsinghua Shenzhen International Graduate School, Tsinghua University, Shenzhen 518055, China} 
\affil[19]{Department of Automation, Tsinghua University, Beijing 100084, China}
\affil[20]{Smart Sensing \& Diagnosis Research Division, Korea Atomic Energy Research Institute, 34057, Republic of Korea}
\affil[21]{Daegu-Gyeongbuk Medical Innovation Foundation, Medical Device Development centre, 427724, Republic of Korea}
\affil[22]{NepAL Applied Mathematics and Informatics Institute for Research (NAAMII), Kathmandu, Nepal}
\affil[23]{Computer Science Department,University of Nottingham, Malaysia Campus, 43500 Semenyih, Malaysia}
\affil[24]{CRAN UMR 7039, Universit\'e de Lorraine and CNRS, F-54500, Vand{\oe}uvre-L{\`es}-Nancy, France}
\affil[25]{Oslo Metropolitan University, Pilestredet 46, 0167 Oslo, Norway}
\affil[26]{Department of Molecular and Clinical Medicine, Sahlgrenska Academy, University of Gothenburg, 41345 Göteborg, Sweden}
\affil[27]{Augere Medical, Nedre Vaskegang 6, 0186 Oslo, Norway}
\affil[*]{corresponding author: Sharib Ali (sharib.ali@eng.ox.ac.uk)}
\affil[$\dag$]{these authors contributed equally to this work}
\keywords{Endoscopy, deep learning, polyp detection, polyp segmentation, generalisability assessment}
\begin{document}
\clearpage
\newpage
\begin{abstract}
Polyps are well-known cancer precursors identified by colonoscopy. However, variability in their size, location, and surface largely affect identification, localisation, and characterisation. Moreover, colonoscopic surveillance and removal of polyps (\textit{referred to as polypectomy}) are highly operator-dependent procedures. 
There exist a high missed detection rate and incomplete removal of colonic polyps due to their variable nature, the difficulties to delineate the abnormality, the high recurrence rates, and the anatomical topography of the colon. There have been several developments in realising automated methods for both detection and segmentation of these polyps using machine learning. However, the major drawback in most of these methods is their ability to generalise to out-of-sample unseen datasets that come from different centres, modalities and acquisition systems. To test this hypothesis rigorously we curated a multi-centre and multi-population dataset acquired from multiple colonoscopy systems and challenged teams comprising machine learning experts to develop robust automated detection and segmentation methods as part of our crowd-sourcing Endoscopic computer vision challenge (EndoCV) 2021. In this paper, we analyse the detection results of the four top (among seven) teams and the segmentation results of the five top teams (among 16). Our analyses demonstrate that the top-ranking teams concentrated on accuracy (i.e., accuracy $> 80\%$ on overall Dice score on different validation sets) over real-time performance required for clinical applicability. We further dissect the methods and provide an experiment-based hypothesis that reveals the need for improved generalisability to tackle diversity present in multi-centre datasets.
\end{abstract}
\flushbottom
\maketitle
\section*{Introduction}
Colorectal cancer (CRC) is the third leading cause of cancer deaths, with reported mortality rate of nearly 51\%~\cite{Bray2018}. CRC can be characterised by early cancer precursors such as adenomas or serrated polyps that may over time lead to cancer. While polypectomy is a standard technique to remove polyps~\cite{kaminski2017increased} by placing a snare (thin wire loop) around the polyp and closing it to cut though the polyp tissue either with diathermy (heat to seal vessels) or without (cold snare polypectomy), identifying small or flat polyps (e.g. lesion less than 10 mm) can be extremely challenging. This is due to complex organ topology of the colon and rectum that make the navigation and treatment procedures difficult and require expert-level skills.   
Similarly, the removal of polyps can be very challenging due to constant organ deformations which make it sometimes impossible to keep track of the lesion boundary making the complete resection difficult and subjective to experience of endoscopists. Computer-assisted systems can help to reduce operator subjectivity and enables  improved adenoma detection rates (ADR). Thus, computer-aided detection and segmentation methods can assist to localise polyps and guide surgical procedures (e.g. polypectomy) by showing the polyp locations and margins. Some of the major requirements of such system to be utilised in clinic are the real-time performance and algorithmic robustness. 

Machine learning, in particular deep learning, together with tremendous improvements in hardware have enabled the possibility to design networks that can provide real-time performance despite their computational complexity. However, one major challenge in developing these methods is the lack of comprehensive public datasets that include diverse patient population, imaging modalities and scope manufacturers. Incorporating real-world challenges in the dataset can only be the way forward in building guaranteed robust systems. In the past, there has been several attempts to collect and curate gastrointestinal (GI) datasets that include other GI lesions and polyps. A summary of existing related datasets with polyps are provided in \textbf{Supplementary Table}~\ref{tab:datasets}. A major limitation of these publicly available datasets is that they consists of either single center or data cohort representing a single population. Additionally, most widely used public datasets have only single frame and single modality images. Moreover, even though conventional white light endoscopy (WLE) is used in standard colonoscopic procedures, narrow-band imaging (NBI), a type of virtual chromo-endoscopy, is widely used by experts for polyp identification and charaterisation. Most deep learning-based detection~\cite{urban2018deep,2019Polyp,zhang2019real} and segmentation~\cite{nguyen2019robust,guo2019giana,zhang2021transfuse,ali2021_endoCV2020} methods are trained and tested on the same center dataset and WLE modality only. These supervised deep learning techniques has a major issue in not being able to generalise to an unseen data from a different center population~\cite{srivastava2021msrf} or even different modality from the same center~\cite{Celik2021:MICCAI}. Also, the type of endoscope used also adds to the compromise in robustness. Due to selective image samples provided by most of the available datasets for method development, the test dataset also comprise of similarly collected set data samples~\cite{silva2014toward,bernal2015wm,borgli2020hyperkvasir,ali2021_endoCV2020}. Similar to the most endoscopic procedures, colonoscopy is a continuous visualisation of mucosa with a camera and a light source. During this process live videos are acquired which are often corrupted with specularity, floating objects, stool, bubbles and pixel saturation~\cite{ali2020objective}. The mucosal scene dynamics such as severe deformations, view-point changes and occlusion can be some major limiting factors for algorithm performance as well. It is thus important to cross examine the generalisability of developed algorithms more comprehensively and on variable data settings including modality changes and continuous frame sequences. 

With the presented Endoscopic Computer Vision (EndoCV) challenge in 2021 we collected and curated a multicenter dataset~\cite{ali2021polypgen} that is aimed at generalisability assessment. For this we took an strategic approach of providing single modality (white light endoscopy modality, WLE) data from five hospitals (both single frame and sequence) for training and validation while the test data consisted of four different configurations - a) mixed center unseen data with WLE modality with samples from centers in training data, b) a different modality data (narrow-band imaging modality, NBI) from all centers, c) a hidden sixth center single frame data and d) a hidden sixth center continuous frame sequence data. While, unseen data with centers included in training assesses the traditional way of testing the supervised machine learning methods on held-out data, unseen modality and hidden center data testing gauge the algorithm's generalisability. Similarly, sequence test data mimics the occurrence of polyps in data as observed in routine clinical colonoscopy procedure.

\subsection*{Detection and localisation}
While classification methods are frame-based classifiers for polyps~\cite{tajbakhsh2015automatic, park2016colonoscopic, ribeiro2016colonic}, detection methods provide both classification and localisation of polyps~\cite{2019Polyp,urban2018deep} which can direct clinicians to the site of interest, and can be additionally used for counting polyps to assess disease burden in patients. With the advancements of object detection architectures, recent methods are end-to-end networks providing better detection performance and improved speed. The state-of-the-art methods are broadly divided into two categories: multi-stage detectors and single-stage detectors. The multi-stage detector methods include Region proposals-Based Convolutional Neural Network (R-CNN)~\cite{girshick2014rich}, Fast R-CNN~\cite{girshick2015fast}, Faster R-CNN~\cite{ren2015faster}, Region-based fully convolutional networks (R-FCN)~\cite{dai2016r}, Feature Pyramid Network (FPN)~\cite{lin2017feature} and Cascade R-CNN~\cite{cai2018cascade}. On the other hand, the One-stage detectors directly provide the predicted output (bounding boxes and object classification) from input images without the region of interest (ROI) proposal stage. The One-stage detector methods include Single-Shot Multibox Detector (SSD)~\cite{liu2016ssd}, Yolo~\cite{redmon2016you}, RetinaNet~\cite{lin2017focal} and Efficientdet~\cite{tan2020efficientdet}.  

Different studies have been conducted in the literature that focused on polyp detection by employing both multi-stage detectors and single-stage detectors. \textbf{Multi-stage Detectors:} Shin et al. \cite{younghak2017} used a transfer learning strategy based on Faster R-CNN architecture with the Inception ResNet backbone to detect polyps. Qadir et al. \cite{2019Polyp} adapted Mask R-CNN~\cite{he2017mask} to detect colorectal polyps and evaluate its performance with different CNN including ResNet50~\cite{he2016deep}, ResNet101~\cite{he2016deep} and Inception ResNetV2~\cite{2016Inception} as its feature extractor. Despite the speed limitation, multi-stage detectors are widely used in the detection task of endoscopy data challenges due to their competitive performance on evaluation metrics. \textbf{Single-stage Detectors:} Urban et al.~\cite{urban2018deep} used YOLO to detect polyps in real-time, which also resulted in high detection performance. Lee et al.~\cite{lee2020real} employed YOLOv2~\cite{redmon2017yolo9000} and validated the proposed approach on four independent dataset. They reported a real-time performance and high sensitivity and specificity on all datasets. Zhang et al.~\cite{zhang2018polyp} proposed the ResYOLO network that adds residual learning modules into the YOLO architecture to train deeper networks. They reported a near-real-time performance for the ResYOLO network depending on the hardware used. Zhang et al.~\cite{zhang2019real} proposed an enhanced SSD named SSD for Gastric Polyps (SSD-GPNet) for real-time gastric polyp detection. SSD-GPNet concatenates feature maps from lower layers and deconvolves higher layers using different pooling techniques. YOLOv3~\cite{redmon2018yolov3} with darknet53 backbone and YOLOv4 showed IOU and average precision (AP) over 0.80\%  and real-time FPS over 45. Moreover, there exist methods that relied on \textbf{anchor-free detectors} to locate the polyps where they claim to detect polyps without the definition of anchors such as CornerNet~\cite{law2018cornernet} and ExtremeNet~\cite{zhou2019bottom}. Zhou et al.~\cite{2019Objects} proposed the CenterNet, which treats each object as a point and increases the speed significantly while ensuring the accuracy is acceptable. While Wang et al.~\cite{wang2019afp} achieved state-of-the-art results on automatic polyp detection in real-time situations using anchor-free object detection methods. In addition to these works, Multi-stage, Single-stage and other types of detectors have been widely used by participants teams in different polyp detection datasets and challenges such as MICCAI'15~\cite{7840040}, ROBUST-MIS~\cite{ross2020robust}, EAD2019~\cite{ali2020objective} and EndoCV2020~\cite{ali2021_endoCV2020}.
\subsection*{Segmentation}
Semantic segmentation is the process of grouping related pixels in an image to an object of the same category. Deep learning has been very successful in the field of the medical domain, convolutional neural networks (CNN) based techniques were suggested to generate complete and precise segmentation outputs without requiring any post-processing. In deep learning, medical segmentation methods can be categorized into four categories: Models based on fully convolutional networks, Models based on Encoder-Decoder architecture, Models based on Pyramid-based architecture and Models based on Dilated Convolution Architecture.

\textbf{Models based on fully convolutional networks:} Brandao et al. \cite{brandao2017fully} proposed three different FCN-based architectures for detection and segmentation of polyps from colonoscopy images. Zhang et al.  \cite{zhang2017automated} proposed multi-step practice for the polyp segmentation. The former step includes region proposal generation using FCN, and the latter step uses spatial features and a random forest classifier for the refinement process. A similar method was introduced by Akbari et al. \cite{akbari2018polyp} which uses patch selection while training FCN and Otsu thresholding to find the accurate location of polyp. Guo et al. \cite{guo2019giana} describe two methods based on FCN for Gastrointestinal ImageANALysis (GIANA) polyp segmentation sub-challenge. 

\textbf{Models based on encoder-decoder architecture:} Nguyen and Slee~\cite{nguyen2019robust} proposed multiple deep encoder-decoder networks to capture multi-level contextual information and learn rich features during training. Zhou et al.~\cite{zhou2019unet++} proposed UNet++, a deeply supervised encoder-decoder network that showed improved performance on polyp segmentation task. Similarly, PraNet~\cite{fan2020pranet} aggregated deep features in their parallel partial decoder to form initial guidance area maps.
Mahmud et al.~\cite{mahmud2021polypsegnet} integrated dilated inception blocks into each unit layer and aggregate the features of the different receptive fields to capture better-generalized feature representations. Huang et al.~\cite{huang2021hardnet} proposed a low memory traffic, fast and accurate method for the polyp segmentation achieving 86 frames per second (FPS). Later, Zhang et al.~\cite{zhang2021transfuse} proposed a hybrid method combining both transformer-based network and CNN to capture global dependencies and the low-level spatial features for the segmentation task. Inspired by high-resolution network~\cite{wang2020deep}, Srivastava et al.~\cite{srivastava2021msrf} proposed multi-scale residual fusion network (MSRF-Net) that allows information exchange across multiple scales and showed improved generalisability on unseen datasets. All of these encoder-decoder architectures were evaluated only on still images. Ji et al.~\cite{ji2021progressively} proposed a progressively normalised self-attention network (PNS-Net) for video polyp segmentation. 

\textbf{Models based on pyramid-based architecture:} Jia et al. \cite{Jia2020} proposed a pyramid-based model named PLPNet for automated pixel-level polyp classification in colonoscopy images. Also, Guo et al. \cite{Guo2019} employed the Pyramid Scene Parsing Network (PSPNet) \cite{Zhao2017} with SegNet \cite{badrinarayanan2017segnet} and U-Net \cite{ronneberger2015u} as an ensemble deep learning model. The proposed model achieved a improvement upto 6.38\% compared with a single basic trainer.

\textbf{Models based on dilated convolution architecture:} Sun et al.~\cite{sun2019colorectal} used dilated convolution in the last block of the encoder while Safarov et al. \cite{safarov2021denseunet} used in all encoder blocks. Though \cite{safarov2021denseunet} used a mesh of attention blocks and residual block as a decoder, both methods tested there model on CVC-ClinicDB achieving F1-score of 96.106 and 96.043, respectively. Furthermore, nested dilation network (NDN) \cite{wang2019nested} was designed to segment lesions and tested on the GIANA2018 dataset achieving improvements on Dice  upto 3\% compared to other methods.
%
\begin{figure}
    \centering
        \includegraphics[width=0.90\textwidth]{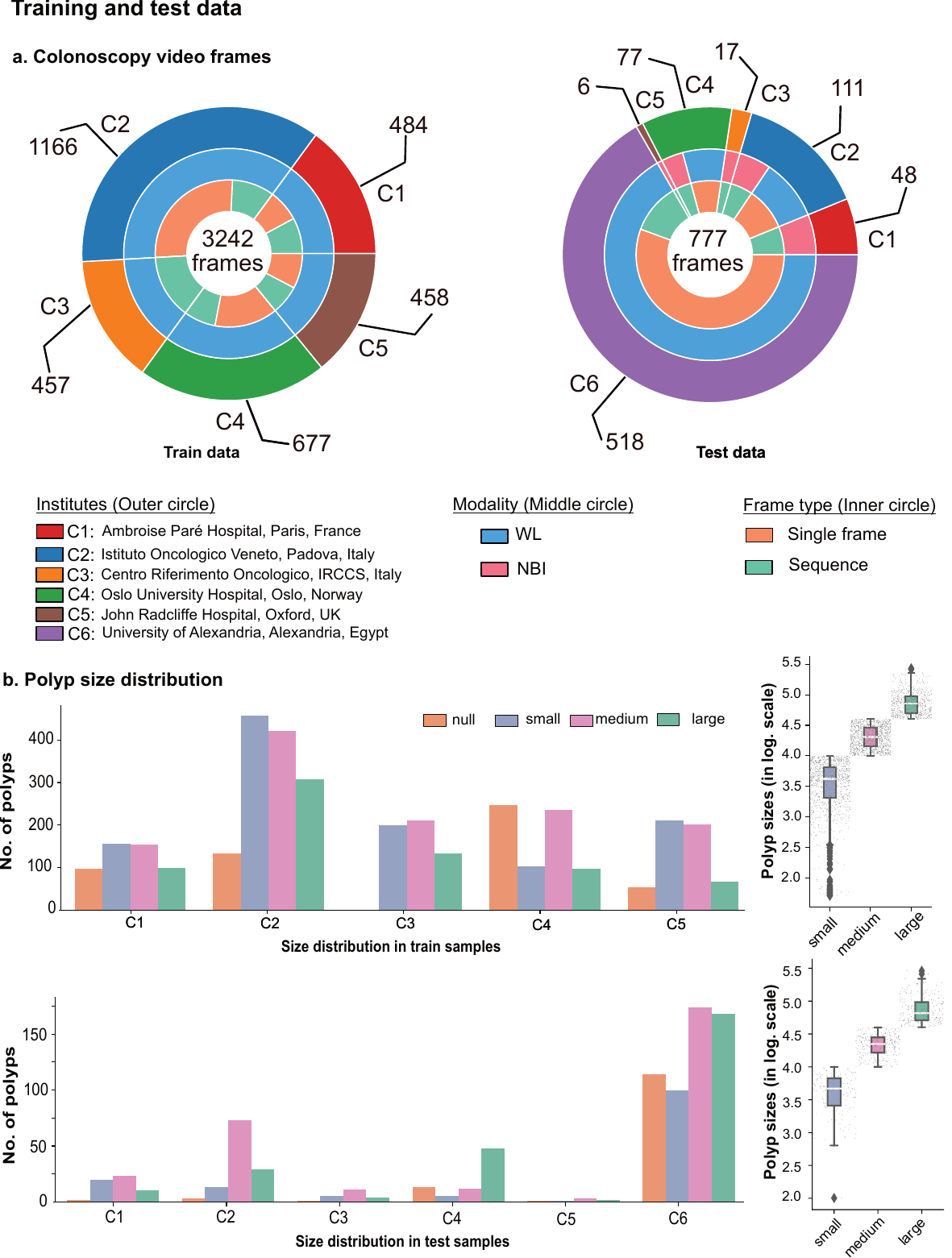}
    \caption{\textbf{Multi-center training and test samples:} a. Colonoscopy video frames for which the annotation samples were reviewed and released as training (left) and test (right) is provided. Training samples included nearly proportional frames from five centers (C1-C5) while test samples consist of majority of single and sequence frames from unseen center (C6); b. Number of polyp or non-polyp samples-based on polyp sizes on resized image frames of $540\times 720$ pixels (left) and their intra-size variability (right) for training (top) and testing data (bottom).}
    \label{fig:my_label_data}
\end{figure}
\begin{figure}[t!]
    \centering
    \includegraphics[width=0.9\textwidth]{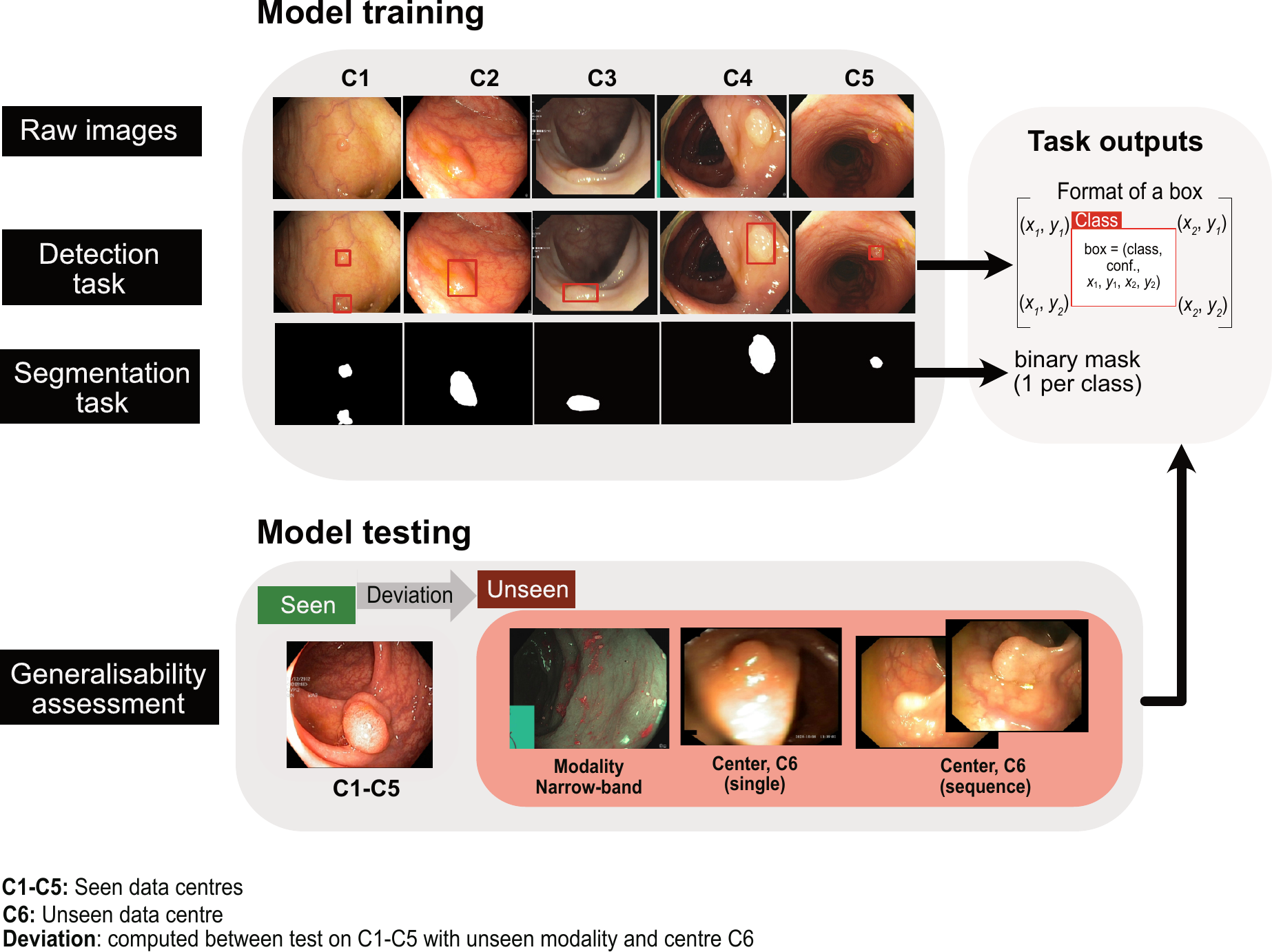}
    \caption{\textbf{EndoCV2021 challenge tasks:} Participants performed model training on white light imaging data collected from five centers (C1-C5). The tasks included detection and segmentation. Trained models were then tested on both seen and unseen center datasets and on unseen data modality (widely used narrow-band imaging). Generalisability assessment is obtained by computing deviations between these unseen samples w.r.t. seen samples. Task outputs included bounding box prediction with confidence and class label for detection task and binary mask prediction for polyp segmentation.}
    \label{fig:challengeTasks}
\end{figure}
\section*{The EndoCV challenge: dataset, evaluation and submission}
In this section, we present the dataset compiled for the polyp generalisation challenge, the protocol used to obtain the ground truth, evaluation metrics that were defined to assess the participants methods and a brief summary on the challenge setup and ranking procedure.
\subsection*{Dataset and challenge tasks}
\paragraph{Dataset:} The EndoCV2021 challenge addresses the generalisability of polyp detection and segmentation tasks in endoscopy frames. The colonoscopy video frames utilised in the challenge are collected from six different centres including two modalities (i.e. WL and NBI) with both sequence and non-sequence frames (see Figure \ref{fig:my_label_data} a). The challenge includes five different types of data and participants were allowed to combine accordingly for their train-validation splits: i) multi-centre video frames from five centres for training and validation, ii) polyp size-based, iii) single frame and sequence data split, iv) modality split (i.e. only for testing phase) and v) one hidden centre test (test phase only). The training dataset consisted of a total of 3242 WL frames from five centers (i.e. C1-C5) with both single and sequence frames. The test dataset, consists of: a) dataset with unseen modality, NBI (data 1), b) dataset with single frames from unknown center (data 2), c) frame sequences from the mixed centers (C1-C5, data 3), and iv) the unseen center sequence frames (C6, data 4). A total of 777 frames were used and data 3 was picked as base dataset against which generalisability of methods were assessed. Polyp size distribution (see Figure~\ref{fig:my_label_data}, left (b)) and its size in log-scale on resized images of the same resolution ($540\times 720$  pixels) (see Figure ~\ref{fig:my_label_data} (b), right) in both training and test sets are presented. These sizes were divided into null (for no polyp in frames), small ($< 100 \times 100$ pixels bounding box), medium ($\leq200\times200$ pixels polyp bounding box) and large ($> 200\times200$ pixels polyp bounding box). These numbers were 534, 1129, 1224 and 705, respectively, for null, small, medium and large size polyps (accounting for 3058 polyp instances) in the training set. Similarly, for the test set the numbers were 134, 144, 296 and 261, respectively, for null, small, medium and large size polyps (in total 701 polyp instances). The size distribution in both of these datasets are nearly identical (see Figure~\ref{fig:my_label_data} (b), right) which is due to the defined range for categorically representing their occurance.


\paragraph{Challenge tasks:} EndoCV2021 included two tasks for which the generalisability assessment was also conducted (see Figure~\ref{fig:challengeTasks}): 1) detection and localisation task and 2) pixel-level segmentation task. For the detection task, participants were provided both single and sequence frames with manually annotated ground truth polyp labels and their corresponding bounding boxes. Participants were required to train their model for predicting class labels, bounding box co-ordinates and confidence scores for localisation. For the segmentation task, we provided the pixel level ground truth segmentation from experts that included the same data as provided for the detection task. Here, the participants were challenged to obtain close to ground truth segmentation binary map prediction. Both of these challenge tasks were assessed rigorously to understand the generalisability of the developed methods. In this regard, the test data consisted of four different categories, here we call it as \textbf{data 1}, \textbf{data 2}, \textbf{data 3} and \textbf{data 4}. Data 1 consisted of unseen modality with NBI data widely used in colonoscopy, data 2 comprises of single frames of unseen center C6, data 3 consisted of mixed seen center (C1-C5) sequence data whereas data 4 included sequence data from unseen center C6. For generalisabilty, we compared the scores between data 3 (seen center data) with the other unseen data categories. All test results were evaluated on a common NVIDIA Tesla V100 GPU. Further details on metric computation is provided in Section \textbf{Evaluation metrics}.
\subsubsection*{Ethical and privacy aspects of the data}
The Data for EndoCV2021 was gathered from 6 different centers located in five different countries  (i.e. UK,  Italy, France, Norway and Egypt). The ethical, legal and privacy of the relevant data was handled by each responsible center. The data collected from each center included a minimum of two essential steps as described below: 
\begin{itemize}
    \item Patient consenting procedure at each institution (required).
    \item Review of the data collection plan by a local medical ethics committee or an institutional review board.
    \item Anonymization of the video or image frames (including demographic information) before sending to the organizers (required).
\end{itemize}

Table \ref{tab:patientConsentingInfo}  illustrates the ethical and legal processes fulfilled by each center along with the details of the endoscopic equipment and recorders used for the collected data. 
\begin{table*}[t!]
{
\begin{tabular}{l|l|l|l|l}
\hline
\textbf{Centers}                              & \textbf{System info.}                & \begin{tabular}[l]{@{}l@{}}\textbf{Ethical}\\ \bf{approval}\end{tabular}  & \begin{tabular}[l]{@{}l@{}}\textbf{Patient consent} \\\textbf{type}\end{tabular}  & \begin{tabular}[l]{@{}l@{}}\textbf{Recording} \\\textbf{system}\end{tabular} \\ \hline
\begin{tabular}[l]{@{}l@{}} Ambroise Par\'{e} Hospital,\\Paris, France    \end{tabular}     & Olympus Exera 195              & \begin{tabular}[l]{@{}l@{}}IDRCB: 2019-\\ A01602-55\end{tabular}      & \begin{tabular}[l]{@{}l@{}}Endospectral\\ study\end{tabular} &  NA  \\ \hline
\begin{tabular}[l]{@{}l@{}} Istituto Oncologico Veneto,\\ Padova, Italy\end{tabular}	     & Olympus endoscope H190 & NA & \begin{tabular}[l]{@{}l@{}}Generic patients \\consent\end{tabular} & \begin{tabular}[l]{@{}l@{}}ENDOX,\\ TESImaging  \end{tabular} \\ \hline
{{\begin{tabular}[l]{@{}l@{}} Centro Riferimento Oncologico,\\ IRCCS, Italy\end{tabular}}}   &\begin{tabular}[l]{@{}l@{}} Olympus VG-165, CV180, \\H185 \end{tabular}         & NA                        & \begin{tabular}[c]{@{}l@{}}Generic patients \\consent\end{tabular}    &   NA       \\ \hline
\begin{tabular}[l]{@{}l@{}}Oslo University Hospital,\\ Oslo, Norway\end{tabular} & \begin{tabular}[l]{@{}l@{}}Olympus Evis Exera III,\\ CF 190 \end{tabular} &Exempted$^\dag$  & \begin{tabular}[l]{@{}l@{}}Written informed\\ consent\end{tabular}   & Pix-E5    \\ \hline
\begin{tabular}[l]{@{}l@{}}John Radcliffe Hospital, \\Oxford, UK\end{tabular}           & \begin{tabular}[l]{@{}l@{}}GIF-H260Z, EVIS Lucera CV260,\\Olympus Medical Systems\end{tabular} & \begin{tabular}[l]{@{}l@{}}REC Ref: \\16/YH/0247 \end{tabular}& Universal consent & MediCapture
\\ \hline
\begin{tabular}[l]{@{}l@{}}University of Alexandria, \\Alexandria, Egypt\end{tabular} & Olympus Exera 160AL, 180AL                             & NA                        & \begin{tabular}[l]{@{}l@{}}Written informed \\consent\end{tabular}  &   NA   \\ \hline
\multicolumn{5}{l}{$^\dag$ \footnotesize{Approved by the data inspectorate. No further ethical approval was required as it did not interfere with patient treatment}}
\end{tabular}
}
\caption{\textbf{Data collection information for each center:} Data acquisition system and patient consenting information. \label{tab:patientConsentingInfo}}
\end{table*}

\subsubsection*{Annotation protocol}
The annotation process was conducted by a team of three experienced researchers using an online annotation tool called Labelbox\footnote{https://labelbox.com}.
Each annotation was cross-validated by the team and by the center expert for accurate boundaries segmentation. At least one senior gastroenterologist was assigned for an independent binary review process. A set of protocols for manual annotation of polyp has been designed as follows: 
\begin{itemize}
 \item Clear raised polyps: Boundary pixels should include only protruded regions. Precaution has to be taken when delineating along the normal colon folds 
\item Inked polyp regions: Only part of the non-inked appearing object delineation 
\item	Polyps with instrument parts: Annotation should not include instrument and is required to be carefully delineated and may form more than one object 
\item	Pedunculated polyps: Annotation should include all raised regions unless appearing on the fold 
\item	Flat polyps: Zooming the regions identified with flat polyps before manual delineation. Also, consulting center expert if needed. 
\end{itemize}

The annotated masks where examined by experienced gastroenterologists who gave a binary score indicating whether a current annotation can be considered clinically acceptable or not. Additionally, some of the experts provided feedback on the annotation where these images were placed into an ambiguous category for further refinement based on the experts feedback. A detailed process along with the number of annotations conducted and reviewed is outline in \textbf{Supplementary Figure}~\ref{fig:suppl_data_annotation}.

\subsection*{Evaluation metrics}
\subsubsection*{Polyp detection}
For the polyp detection task, we have computed standard computer vision metrics such as average precision (AP) and intersection-of-union (IoU)~\cite{lin2014microsoft}. 

\begin{itemize}
\item {IoU}: The IoU metrics measures the overlap between two bounding boxes A and B as the ratio between the target mask and predicted output.
\begin{equation}
\text{IoU(A,B)} =\frac{A \cap B} {A \cup B} 
\end{equation}
Here, $\cap$ represents intersection and $\cup$ represents the union. \item {AP}: AP is computed as the Area Under Curve (AUC) of the precision-recall curve of detection sampled at all unique recall values (r1, r2, ...) whenever the maximum precision value drops:
\begin{equation}
    \mathrm{AP} = \sum_n{\left\{\left(r_{n+1}-r_{n}\right)p_{\mathrm{interp}}(r_{n+1})\right\}}, 
\end{equation}
with $p_{\mathrm{interp}}(r_{n+1}) =\underset{\tilde{r}\ge r_{n+1}}{\max}p(\tilde{r})$. Here, $p(r_n)$ denotes the precision value at a given recall value. This definition ensures monotonically decreasing precision. AP was computed as an average APs at $0.50$  and $0.95$ with the increment of 0.05. Additionally, we have calculated
AP\textsubscript{\textit{small}}, AP\textsubscript{\textit{medium}}, AP\textsubscript{\textit{large}}.  More description about the detection evaluation metrics and their formulas can be found here~\footnote{{https://github.com/sharibox/EndoCV2021-polyp\_det\_seg\_gen/blob/main/evaluationMetrics/coco\_evaluator.py}}. 
\end{itemize}
\subsubsection*{Polyp segmentation }
For polyp segmentation task, we have used widely accepted computer vision metrics that include Sørensen–Dice Coefficient ($DSC = \frac{2 \cdot tp} {2 \cdot tp + fp + fn}$), {Jaccard Coefficient} ($JC= \frac{tp} {tp + fp + fn}$), precision ($p=\frac{tp} {tp + fp}$), and recall ($r= \frac{tp} {tp + fn}$), overall accuracy \textit{($Acc = {\frac{tp + tn} {tp + tn + fp + fn}}$ )}, and F2 ($=\frac{5p \times r} {4p + r}$). In addition to the performance metrics, we have also computed frame per second (FPS$ = {\frac{\#frames} {sec}}$). Here,  \textit{tp}, \textit{fp}, \textit{tn}, and fn represent true positives, false positives, true negatives, and false negatives, respectively. 

Another commonly used segmentation metric that is based on the distance between two point sets, here ground truth (G) and estimated or predicted (E) pixels, to estimate ranking errors is the average Hausdorff distance ($H_{d}$) and defined as:
\begin{equation}
    H_{d}(G, E) = \bigg(\frac{1}{G} \sum_{g\in G} \min_{e\in E} d(g, e) + \frac{1}{E} \sum_{e\in E} \min_{g\in G} d (g, e)\bigg)/2.
\end{equation}

The mean ${H}_{d}$ is normalised between 0 and 1 by dividing it by the maximum value $H_{{d}_{max}}$ for a given test set.
\subsubsection*{Polyp generalisation metrics}
We define the generalisability score based on the stability of the algorithm performance on seen white light modality and center dataset (data 3) with unseen center split (data 2 and data 4) and unseen modality (data 1) in the test dataset. We conducted the generalisability assessment for both detection and segmentation approaches separately. 

For detection, the deviation in score between seen and unseen data types are computed over different $AP$ categories, $k \in \{mean, small, medium, large\}$ with tolerance, ($tl = 0.1$):

\begin{equation}
    dev\_g = \frac{1}{|k|}\sum_k\begin{cases}
    |\mathrm{AP_k}^{seen} - \mathrm{AP_k}^{unseen}|, & \text{if } \mathrm{AP_k}^{seen} - tl *\mathrm{AP_k}^{seen} \leq  \mathrm{AP_k}^{unseen} \geq \mathrm{AP_k}^{seen} + tl *\mathrm{AP_k}^{seen}\\
                            0, \quad \text{otherwise.} \\
                         \end{cases}
\end{equation}

Similarly, for segmentation, the deviation in score between seen and unseen data types are computed over different segmentation metric categories, $k \in \{DSC, F2, p, r, H_d\}$ with tolerance, ($tl = 0.05$):
\begin{equation}
    dev\_g = \frac{1}{|k|}\sum_k\begin{cases}
    |\mathrm{S_k}^{seen}  - \mathrm{S_k}^{unseen}|, & \text{for } \mathrm{S_k}^{seen} - tl *\mathrm{S_k}^{seen} \leq  \mathrm{S_k}^{unseen} \geq \mathrm{S_k}^{seen} + tl *\mathrm{S_k}^{seen}\\
                            0, \quad \text{otherwise.} \\
                         \end{cases}
\end{equation}
\subsection*{Challenge setup, and ranking procedure}
We set-up challenge website\footnote{https://endocv2021.grand-challenge.org} with an automated docker system for metric-based ranking procedure. Challenge participants were required to perform inference on our cloud-based system that incorporated NVIDIA Tesla V100 GPU and provided test dataset with instructions of using directly on GPU without downloading the data for round 1 and round 2. However, we added an additional round 3 where the challenge participant's trained model were used for inference by the organisers on an additional unseen sequence dataset. Thus, the challenge consisted of three rounds. All provided test frames were from unseen patient data to avoid any data leakage. Further details on data samples in each round are summarised below:
\begin{itemize}
    \item {\bf{Round 1:}} Test subset-I consisted of: a) 50 samples of each data 1 (unseen modality), data 2 (unseen single sample, C6) and data 3 (mixed center C1-C5 sequence data)
    \item {\bf{Round 2:}}  Test subset-II comprised of: a) all 88 samples of each data 1 (unseen modality), 86 samples of data 2 (unseen single sample, C6) and 124 samples of data 3 (mixed C1-C5)
    \item {\bf{Round 3:}} Inference on round 3 data was performed by the organisers on the same GPU. This round comprised of: a) 135 samples of each data 1 (unseen modality), 86 samples of data 2 (unseen single sample, C6), 124 samples of data 3 (mixed center C1-C5 sequence data) and an additional set of 432 sequence samples from unseen center C6. 
\end{itemize}

We conducted elimination for both round 1 and round 2 which was based on the metric scores on the leaderboard and timely submission. In round 2, we eliminated those with very high computational time for inference and metric score consistency. The chosen participants were requested for the method description paper at the EndoCV proceeding~\cite{Ali2021:EndoCVProc} to allow transparent reporting of their methods. All accepted methods were eligible for round 3 evaluation and have been reported in this paper.

Detection ranking was performed as an aggregated score between the average precision and deviation scores between data 3 (seen C1-C5) w.r.t. other unseen data in the test set. Similarly, for team ranking on segmentation, we used segmentation metrics and deviation scores for segmentation (between seen and unseen data). Please refer to Section Evaluation metrics for details. An aggregated rank was used to announce winner. In our final ranking reported in this paper, we have additionally used inference time as well. 
\section*{Method summary of the participants}
Below, we summarise the EndoCV2021 generalisability assessment challenge for polyp detection and segmentation methods using deep learning. Tabulated summaries are also provided highlighting the nature of the devised methods and basis of choice in-terms of speed and accuracy for detection (see Table~\ref{table:challenge_summary_detection}) and segmentation (see Table~\ref{table:challenge_summary_segmentation}). Methods are detailed in the compiled EndoCV2021 challenge proceeding~\cite{Ali2021:EndoCVProc}. 
\subsection*{Detection Task}
\begin{table*} [!t]
\centering
\caption{{Summary of the participating teams \textbf{detection task} for EndoCV2021 Challenge}. All test was done on NVIDIA V100 GPU provided by the organisers.}
\begin{tabular}{lllllllll}
\toprule
\bf Team Name &\bf Algorithm & \bf Backbone &\bf Nature  &\bf \begin{tabular}[l]{@{}c@{}}
    Choice  \\ Basis\end{tabular} & \bf \begin{tabular}[l]{@{}c@{}}
    Data  \\ Aug.\end{tabular} & \bf 
    Loss  & \bf Opt.&\bf Code\\  \midrule  \midrule
AIM\_CityU \cite{lia2021joint} & FCOS & \begin{tabular}[l]{@{}l@{}}FPN, ResNeXt\\ -101-DCN\end{tabular}  & ATSS & \begin{tabular}[l]{@{}c@{}}
    Accuracy  \\
      speed 
\end{tabular}  & No &  \begin{tabular}[l]{@{}l@{}}Generalised \\Focal Loss\end{tabular}  & SGD  &\href{https://github.com/CityU-AIM-Group/EndoCV-2021}{\nolinkurl{[d1]}} \\  
HoLLYS\_ETRI \cite{honga2021deep} & Mask R-CNN  &  \begin{tabular}[l]{@{}l@{}}FPN\\ ResNet34\end{tabular}   & Ensemble & Accuracy++  & No & Smooth L1   & SGD & \href{https://github.com/EndoCV2021/detectron}{\nolinkurl{[d2]}} \\
JIN\_ZJU \cite{gana2021detection}  & YOLOV5 &\begin{tabular}[l]{@{}l@{}}CSPdarknet\\ SPP\end{tabular}  & Ensemble & speed++ & Yes & BECLogits & SGD & \href{https://github.com/GTYuantt/EndoCV2021\_yolov5}{\nolinkurl{[d3]}} \\ 
GECE\_VISION \cite{polat2021polyp} & EfficientDet \cite{tan2020efficientdet}  & \begin{tabular}[l]{@{}l@{}}EfficientNet \\D0-D3\end{tabular}   & Ensemble & Accuracy & Yes & - & Adam &\href{https://github.com/GorkemP/EndoCV2021-EfficientDet-Pytorch}{\nolinkurl{[d4]}} \\ 
\bottomrule
\multicolumn{9}{l}{\footnotesize{FCOS: Fully Convolutional One-Stage Object Detection; FPN: Feature Pyramid Network; ATSS: Adaptive Training Sample Selection}}\\
\multicolumn{9}{l}{\footnotesize{YOLO: You Only Look Once; SGD: Stochastic Gradient Descent; [d1]-[d4]: hyperlinked GitHub repos.}}
\end{tabular}
 \label{table:challenge_summary_detection}
\end{table*}

\begin{itemize}
    \item \textbf{AIM\_CityU:}~\cite{lia2021joint} The team used one-stage anchor-free FCOS~\cite{tian2019fcos} as the baseline detection algorithm and adopted ResNeXt-101-DCN with FPN as their final feature extractor. For the model optimization, both online (random flipping and multi-scale training) and offline (random rotation, gamma contrast, and brightness transformation, etc.) data augmentation strategies are performed to improve the model generalization. 
    \item \textbf{HoLLYS\underline{ }ETRI:}~\cite{honga2021deep}
    The team used Mask R-CNN~\cite{he2017mask} for detection and segmentation task. All the weights were initialized with pre-trained weights. An ensemble learning method based on 5-fold cross-validation was used to improve the generalization performance. While training a single Mask R-CNN, the data acquired from all data centers were not used. Instead, only the data acquired from four centers were used for training and, the data from remained center was used for validation. 
    Ensemble inference was performed by combining the inference results of 5 models. For the detection task, weighted box fusion technique~\cite{solovyev2021weighted} was used to combining results of detection. For segmentation task, segmentation masks from 5 models were averaged with IoU threshold of 0.6.
    
    \item \textbf{JIN\_ZJU:}~\cite{gana2021detection} The team used the YoloV5~\cite{yolov5} as the baseline detection algorithm. To improve the generalisation ability of the standard Yolov5, different data augmentation methods were applied that included hue adjustment, saturation adjustment, value adjustment, rotating, translation, scaling, up-down flipping, left-right flipping, mosaic and mixup.
    
    \item \textbf{GECE\_VISION:} \cite{polat2021polyp}
    The team proposed an ensemble-based polyp detection architecture using the EfficientDet \cite{tan2020efficientdet} as the base model family with EfficientNet as backbone network.  The bootstrap aggregating (bagging) was utilized to aggregate different versions of the predictors (EfficientDet D0, D1, D2, D3) which are trained on bootstrap replicates of the training set.
    In order to increase the variance and improve generalization capability of the model, data augmentation have been used (i.e., scale jittering with 0.2-2.0, horizontal flipping, and rotating between 0$^\circ$-360$^\circ$). Adam optimiser and the scheduling learning rate were used with decreasing factor of 0.2 whenever validation loss did not change in the last 10 epochs. 
 \end{itemize}
\subsection*{Segmentation Task}
%
\begin{table*} [t!]
\centering
\caption{{Summary of the participating teams algorithm for \textbf{segmentation task} EndoCV2021 Challenge}. Top five teams are shown above horizontal line and worse performing team in round 3 are provided below this line.}
\begin{tabular}{lllllllll}
\toprule
\bf Team Name &\bf Algorithm & \bf Backbone &\bf Nature  &\bf \begin{tabular}[l]{@{}c@{}}
    Choice  \\ Basis\end{tabular} & \bf \begin{tabular}[l]{@{}c@{}}
    Data  \\ Aug.\end{tabular} & \bf \begin{tabular}[l]{@{}c@{}}
    Loss  \\ function\end{tabular}& \bf Optimizer &\bf Code\\  \midrule  \midrule
aggcmab \cite{ galdrana2021multi} & DPN92-FPN & DPN92-FPN  & Cascaded & Accuracy++ & Yes &BCE  & SGD& \href{https://github.com/agaldran/endocv21}{\nolinkurl{[s1]}}\\
AIM\_CityU~\cite{lia2021joint} & HRNet + LRM & HRNet  & MSFF & \begin{tabular}[l]{@{}l@{}}Accuracy\\speed\end{tabular}  & Yes &  \begin{tabular}[l]{@{}l@{}}BCE,\\DSC\end{tabular}  & SGD & \href{https://github.com/CityU-AIM-Group/EndoCV-2021}{\nolinkurl{[s2]}} \\
HoLLYS\_ETRI \cite{honga2021deep}& Mask R-CNN & ResNet50  & Ensemble & \begin{tabular}[l]{@{}l@{}}Accuracy+\\speed+\end{tabular} & Yes & \begin{tabular}[l]{@{}l@{}}Smooth\\ L1\end{tabular}   & SGD  & 
\href{https://github.com/EndoCV2021/detectron }{\nolinkurl{[s3]}}
\\
MLC\_SimulaMet~\cite{thambawita2021divergentnets} &DivergentNet  &  TriUNet  & Ensemble & Accuracy++ & No & \begin{tabular}[l]{@{}c@{}}
    BCE, \\DSC 
\end{tabular}& Adam & \href{https://github.com/vlbthambawita/divergent-nets}{\nolinkurl{[s4]}}
\\ 
sruniga \cite{ghimirea2021augmentation} & HarDNet68  & HarDNet68  &Multiscale & \begin{tabular}[l]{@{}l@{}}Accuracy+\\speed++\end{tabular}& No & BCE & Adam& \href{https://github.com/SahadevPoudel/HarDNet-MSEG}{\nolinkurl{[s5]}}\\ 
%
\hline
YCH\_THU \cite{yua2021parallel} & PraNet  &  Res2Net   & \begin{tabular}[l]{@{}l@{}}Reverse\\ attention\end{tabular}  &\begin{tabular}[l]{@{}l@{}}Accuracy\\speed\end{tabular}  & No &  \begin{tabular}[l]{@{}l@{}}W. IOU\\ W. BCE\end{tabular}  &Adam  & 
\href{https://github.com/kirtsy/Polyp-Segmentation--endoCV2021}{\nolinkurl{[s6]}}
\\ 
Mah\_UNM \cite{haithamia2021embedded} & SegNet & VGG16  & GRU & \begin{tabular}[l]{@{}l@{}}Accuracy\\speed\end{tabular} & No & BCE & Adam&  
\href{https://github.com/mss3331/EndoCV21_SegNetGRU}{\nolinkurl{[s7]}}
\\ 
NDS\_MultiUni \cite{tomara2021improving} &MultiResUnet  & ResUnet  & Ensemble& \begin{tabular}[l]{@{}l@{}}Accuracy\end{tabular}  & No & BCE & Adam & NA \\ 
\bottomrule
\multicolumn{9}{l}{\footnotesize{LRM: Low-rank module; MSFF: Multi-scale feature fusion; DPN: Dual path network; FPN: Feature pyramid network; BCE: Binary cross entropy}}\\
\multicolumn{9}{l}{\footnotesize{BCE: Binary cross entropy; DSC: Dice similarity coefficient; IoU: Intersection over union; W: weighted; SGD: Stochastic gradient descent}}
\end{tabular}
\label{table:challenge_summary_segmentation}
\end{table*}

\begin{itemize}
\item \textbf{aggcmab:} The team~\cite{ galdrana2021multi} improved their previously developed framework cascaded double encoder-decoder convolutional neural network~\cite{galdran2021double} by increasing the encoder representation capability and adapting to a multi-site sampling technique. The first encoder-decoder generates an initial attempt to segment the polyp by extracting features and downsampling spatial resolutions while increasing the number of channels by learning convolutional filters. The output from the first network acts as an input for the second encoder-decoder along with the original image.
Cross-entropy loss was minimized using the stochastic gradient descent with a batch-size of 4 and a learning rate of \textit{lr} = 0.01 with rate decay of 1{e–8} every 25 epochs. The training images were resized to 640$\times$512, and data augmentation (e.g. random rotations, vertical/horizontal flipping, contrast, saturation and brightness changes) was applied.
Four versions were generated from the image (i.e. horizontal and vertical flipping), and the average result was calculated on the test set. 

\item \textbf{AIM\_CityU:} The team~\cite{lia2021joint} adopted HRNet~\cite{wang2020deep} as the backbone to maintain the high-resolution representations in multi-scale feature fusion mechanism. To further eliminate noisy information in segmentation predictions and enhance model generalization, the team proposed a low-rank module to distribute feature maps in the high dimensional space to a low dimensional manifold. For the model optimization, various data augmentation strategies, including random flipping, rotation, color shift (brightness, color, sharpness, and contrast) and Gaussian noise, were performed to improve the model generalization further. Cross entropy and dice loss are utilized to optimize the whole model.
 
\item \textbf{HoLLYS\_ETRI:} The team~\cite{honga2021deep} proposed an ensemble inference model based on 5-fold cross-validation to improve the performance of polyp detection and segmentation. The Mask R-CNN was used to generate the output segmentation mask. Ensemble inference was used to generate the final segmentation mask by averaging the results from the 5 models. After averaging the masks, if the inference results were greater than the threshold (0.6) then the output mask is considered as a polyp otherwise was counted as a background. Data augmentation was performed based on the techniques provided in Detectron2. The model was trained for 50,000 steps and checkpoints were save for every 1,000 steps with learning rate \textit{lr}=0.001 that changes with a warm-up scheduler
 \item \textbf{MLC\_SimulaMet:} The team~\cite{thambawita2021divergentnets} developed two ensemble models using well-known segmentation models; namely UNet++~\cite{zhou2019unet++}, FPN~\cite{lin2017feature}, DeepLabv3~\cite{chen2017rethinking}, DeepLabv3+~\cite{chen2018encoder} and novel TriUNet for their DivergentNet ensemble model, and three UNet~\cite{ronneberger2015u} architectures in their TriUNet ensemble model. The novel TriUNet model takes a single image as input, which is passed through two separate UNet models with different randomized weights. The output of both models was then concatenated before being passed through a third UNet model to predict the final segmentation mask. The whole TriUNet network was trained as a single unit. Thus, the proposed DivergentNet included five segmentation models.
\item \textbf{sruniga:} 
The team~\cite{ghimirea2021augmentation} suggested a lightweight deep learning-based algorithm to meet the real-time clinical need. The proposed network applied the HarDNet-MSEG~\cite{huang2021hardnet} as the backbone network as it has a low inference speed due to reduced shortcuts. Moreover, they proposed an augmentation strategy for realising improved generalizable model.
The data augmentation was applied according to a certain probability.  
For training the model, the dataset was split into 80\% training and 20\% validation using adam optimizer and setting the learning rate \textit{lr} of 0.00001 for all the experiments. 
Images were resized to 352$\times$352, and data augmentation has been applied according to the proposed algorithm by the team.  
\item \textbf{Mah\_UNM:}
The team~\cite{haithamia2021embedded} proposed modifying the SegNet~\cite{badrinarayanan2017segnet} by embedding Gated recurrent units (GRU) units~\cite{cho2014learning } within the convolution layers to improve its performance in segmenting polyps. 
The hyperparameters were set as the original SegNet with learning rate \textit{lr} of 0.005 and batch size of 4. The multiplicative factor of gamma of 0.8 was used for the learning rate decay with adam optimizer and weighted cross-entropy loss. The provided dataset was split into 80\% training and 20\% validation.  


\item \textbf{NDS\_MultiUni:} The team~\cite{tomara2021improving} suggested building a cascaded ensemble model made of MultiResUNet \cite{ibtehaz2020multiresunet} architectures. The input image was fed to four different MultiResUNet models in the proposed model, and each model generated an output mask. Afterwards, the four predicted outputs were averaged together to produce the final segmentation mask. Each model was trained for 100 epochs with the same setting of hyperparameters. The input images were resized to 256$\times$256 with a batch size of 8, binary cross-entropy as loss function and using Adam optimizer. The learning rate is set to \textit{lr} of $1{e}-3$ and using the Reduce LROnPlateau callback. 
   
\item \textbf{YCH\_THU:} The team~\cite{yua2021parallel} used existing parallel reverse attention network (PraNet) \cite{fan2020pranet}. They extracted multi-level features from colonoscopy images utilizing a parallel res2Net-based network. Moreover, the segmentation results are post-processed to remove uncertain pixels and enhance the boundary. The images were resized to 512$\times$512 and the dataset was split into 80\% training and 20\% validation. The model was trained for 300 epochs with batch size 18 and learning rate \textit{lr} of 1\textit{e}-4 which was reduced every 50 epochs. 
\end{itemize}
\section*{Results}
The EndoCV2021 challenge focus on detection and segmentation of polyps with different sizes from endoscopic frames. The endoscopy video frames are gathered from six worldwide centers  including two different modalities (i.e. White Light and Narrow Band Imaging). The frames were annotated by clinical experts in the challenge team for the purpose of detection and localization. The training dataset consisted of total 3242 frames from five centers only with the release of binary masks for the segmentation task and bounding box coordinates for detection task. For the test dataset,  frames from center six was include to provide an overall of 777 frames from the six centers with a variation between single and sequence frames. There was a variation in the polyp size in both the training and testing set as shown in Fig.~\ref{fig:my_label_data}b. 
%
\subsection*{Aggregated performance and ranking on detection task}
Table \ref{detection_table_1} represents the average precision (AP) computed at three different IoU thresholds and AP at different scales for the participant teams on the four datasets. Moreover, results from baseline methods YOLOv4~\cite{YOLOv4-Alex2020}, RetinaNet (ResNet50) \cite{lin2017focal} and EfficientNet-D2 are provided. Methods presented by teams \textit{ HoLLYS\_ETRI } and \textit{ JIN\_ZJU } outperform against the other teams in terms of AP values for the single frame datasets (i.e. both data 1 (NBI) and data 2 (WLE). The results by both teams on data 1 had an increased difference for AP$_{mean}$ (>20\%), AP$_{50}$ (>15\%) and AP$_{75}$ (>18\%) when compared to the other teams.  However, for data 2, team \textit{AIM\_CityU} produced comparable results leading them to third place with a small difference of 0.88\% for AP$_{mean}$ score when compared to team \textit{ HoLLYS\_ETRI }. 

For the seen sequence dataset (Data 3), team \textit{ JIN\_ZJU} maintained the top performance for AP$_{mean}$ (i.e. higher than second-best team \textit{AIM\_CityU} by 4.19\%) and AP{$_{75}$} (i.e. higher than second-best team \textit{HoLLYS\_ETRI} by 3.29\%). Team \textit{HoLLYS\_ETRI} maintained their top performance with highest result for AP{$_{50}$} with a greater difference of 2.10\% when compared to \textit{AIM\_CityU} that comes in second place.  Furthermore, the method by \textit{ HoLLYS\_ETRI} surpassed the results of other teams and baseline methods on the unseen sequence (Data 4) where the second teams take place with a difference of (>0.037) AP$_{mean}$, (>0.04) AP{$_{50}$} and (>0.055) AP{$_{75}$}. In general, as concluded from table \ref{detection_table_1}, results by  teams \textit{HoLLYS\_ETRI}, \textit{JIN\_ZJU} and \textit{AIM\_CityU} had the best performance while results of the baselines method did not show better performance compared to any proposed method.

Table \ref{detection_table_2} shows the ranking of detection task of polyp generalisation challenge after calculating the average detection, average deviation scores and time. Team \textit{AIM\_CityU} ranks the first place with inference time of 0.10 second per frame and lowest deviation scores of dev\_g$_{2-3}$ (0.1339),  dev\_g$_{4-3}$ (0.0562) and dev\_g (0.0932). Followed by team \textit{HoLLYS\_ETRI} in second place with an increased inference time of 0.69 s per frame and a top score of 0.4914 for average detection. In third place, team \textit{JIN\_ZJU} takes place with 1.9 s per frame for the inference time and the second-best average detection result of 0.4783. 
\begin{table}[t!]
\centering
\caption{Team results for the {detection task} with average precision AP computed at IoU thresholds 50 (AP$_{50}$), 75 (AP$_{75}$), and $[0.50:0.05:0.95]$ mean AP (AP$_{mean}$). Size wise AP values are also presented.}
\small
\begin{tabular}{l|l|l|ll|lll}
\toprule
 &  & \multicolumn{3}{c|}{\bf{Average precision, AP}} &  \multicolumn{3}{|c}{{\bf AP across scales}} \\
  \cline{3-8}
{\bf{Data type}} & \bf{Teams/Method} & \multicolumn{3}{|c|}{} &\multicolumn{3}{|c}{} \\
 & & \bf{AP}$_{mean}$      & \bf{AP}{$_{50}$}   & \bf{AP}{$_{75}$}   & \multicolumn{1}{|c}{}\bf{AP}$_{small}$ & \bf{AP}$_{medium}$ & \bf{AP}$_{large}$ \\ \hline \midrule  
\multirow{6}{*}{\rotatebox[origin=c]{90}{\textbf{{{\begin{tabular}[l]{@{}c@{}}Data 1 \\ (NBI-single)  \end{tabular}}}}}} 
& AIM\_CityU~\cite{lia2021joint}& 0.351  & 0.5378 & 0.3988 & \textbf{0.0802}  & \textbf{0.3213}   & 0.4076   \\
& GECE\_VISION~\cite{polat2021polyp}         & 0.3182 & 0.5266 & 0.3498 & 0.0506  & 0.1863   & 0.3984   \\
& HoLLYS\_ETRI \cite{honga2021deep}              & \textbf{0.4743} & \textbf{0.6931} & \textbf{0.5517 }& \textbf{0.13}    & \textbf{0.3961}   & \textbf{0.5502}   \\
& JIN\_ZJU~\cite{gana2021detection}  & \textbf{0.4461} & \textbf{0.6587} & \textbf{0.498}  & 0.0378  & 0.2963   & \textbf{0.5866}   \\
\cline{2-8}
& YOLOv4~\cite{YOLOv4-Alex2020}             & 0.3099 & 0.4472 & 0.3719 & 0.0688  & 0.2546   & 0.3711   \\
& \multicolumn{1}{l|}{{\begin{tabular}[l]{@{}c@{}}RetinaNet (ResNet50) \cite{lin2017focal}  \end{tabular}}}      & 0.3145  & 0.5625	& 0.2673 &	0.0475 &	0.223 &	0.3805 	\\
& \multicolumn{1}{l|}{{\begin{tabular}[l]{@{}c@{}}EfficientNet-D2 \cite{tan2020efficientdet} \end{tabular}}}  & 0.2009 &	0.3092 &	0.2279	& 0.0297 &	0.1591	& 0.2409 	\\
\bottomrule 
\multirow{6}{*}{\rotatebox[origin=c]{90}{\textbf{{{\begin{tabular}[l]{@{}c@{}}Data 2 \\ (WLE-single)  \end{tabular}}}}}} 
& AIM\_CityU~\cite{lia2021joint}           & 0.5733 & 0.7847 & 0.6058 & 0.2799  & 0.4835   & \textbf{0.6595}   \\
& GECE\_VISION~\cite{polat2021polyp}           & 0.5327 & 0.7859 & 0.5357 & 0.1556  & 0.4595   & 0.6235   \\
& HoLLYS\_ETRI \cite{honga2021deep}   & \textbf{0.5784} & \textbf{0.7908} & \textbf{0.6808} & \textbf{0.3854 } & \textbf{0.4981}   & 0.6552   \\
& JIN\_ZJU~\cite{gana2021detection} & \textbf{0.6049} & \textbf{0.8095} & \textbf{0.6643} & \textbf{0.3071}  & \textbf{0.6148}   & \textbf{0.721}    \\
\cline{2-8}
& YOLOv4~\cite{YOLOv4-Alex2020}               & 0.4194 & 0.5996 & 0.4636 & 0       & 0.3337   & 0.5237   \\
& \multicolumn{1}{l|}{{\begin{tabular}[c]{@{}c@{}}RetinaNet \cite{lin2017focal} (ResNet50)  \end{tabular}}}    &    0.4076 & 	0.7355 & 	0.4607	 & 0 & 	0.2743 & 	0.5246
\\
& \multicolumn{1}{l|}{{\begin{tabular}[l]{@{}c@{}}EfficientNet-D2 \cite{tan2020efficientdet} \end{tabular}}}  & 0.4204 & 	0.6135 & 	0.4647 & 	0 & 	0.3828	 & 0.5127	\\
\bottomrule 
\multirow{6}{*}{\rotatebox[origin=c]{90}{\textbf{{{\begin{tabular}[l]{@{}c@{}}Data 3 \\ (seen seq.)  \end{tabular}}}}}} 
& AIM\_CityU~\cite{lia2021joint}           & \textbf{0.5296} & \textbf{0.7804} & 0.5484 & \textbf{0.0035} & \textbf{0.404}    & 0.5784   \\
& GECE\_VISION~\cite{polat2021polyp}           & 0.3725 & 0.6589 & 0.3844 & 0.005   & 0.0269   & 0.4363   \\
& HoLLYS\_ETRI \cite{honga2021deep}             & 0.5287 & \textbf{0.7972} & \textbf{0.5759} & \textbf{0.0174}  & 0.0236   & \textbf{0.599}   \\
& JIN\_ZJU~\cite{gana2021detection}              & \textbf{0.5528} & 0.7253 & \textbf{0.5955} & 0       & \textbf{0.1515}   & \textbf{0.6515}   \\
\cline{2-8}
& YOLOv4~\cite{YOLOv4-Alex2020}               & 0.2987 & 0.4362 & 0.3544 & 0       & 0        & 0.3283   \\
& \multicolumn{1}{l|}{{\begin{tabular}[c]{@{}c@{}}RetinaNet (ResNet50) \cite{lin2017focal}  \end{tabular}}}  & 0.3119 & 	0.4897	 & 0.3564 & 	0 & 	0.2525 & 	0.341
\\
& \multicolumn{1}{l|}{{\begin{tabular}[l]{@{}c@{}}EfficientNet-D2 \cite{tan2020efficientdet} \end{tabular}}}  & 0.2933	&  0.4035 & 	0.3758 & 	0 & 	0.0757	 & 0.3238 	\\
\bottomrule 
\multirow{6}{*}{\rotatebox[origin=c]{90}{\textbf{{{\begin{tabular}[l]{@{}c@{}}Data 4 \\ (unseen seq.)  \end{tabular}}}}}} 
& AIM\_CityU~\cite{lia2021joint}   & \textbf{0.3464} & 0.4725 & \textbf{0.3767} & 0       & 0.2723   & 0.5225   \\
& GECE\_VISION~\cite{polat2021polyp}           & 0.3146 & \textbf{0.4997} & 0.3302 & 0       & 0.2787   & 0.4718   \\
& HoLLYS\_ETRI \cite{honga2021deep}   & \textbf{0.3843} & \textbf{0.5402} & \textbf{0.4318} & \textbf{0.0002} & \textbf{0.3096}   & \textbf{0.5802}   \\
& JIN\_ZJU~\cite{gana2021detection}             & 0.3094 & 0.4259 & 0.3301 & \textbf{0.0001}  & \textbf{0.3154}  & \textbf{0.6568}   \\
\cline{2-8}
& YOLOv4~\cite{YOLOv4-Alex2020}               & 0.2363 & 0.3105 & 0.2805 & 0       & 0.2083   & 0.3498   \\
& \multicolumn{1}{l|}{{\begin{tabular}[c]{@{}c@{}}RetinaNet (ResNet50) \cite{lin2017focal}  \end{tabular}}}       & 0.2487	&  0.4491 & 	0.2657 &  	0.001 & 	0.1763	&  0.385
\\
& \multicolumn{1}{l|}{{\begin{tabular}[l]{@{}c@{}}EfficientNet-D2 \cite{tan2020efficientdet} \end{tabular}}}  & 0.2787 & 	0.3818 & 	0.3365 & 	0 & 	0.2647	 &  0.4078	\\
\bottomrule 
\end{tabular}
\label{detection_table_1}
\end{table}
\begin{figure}
    \centering
    \includegraphics{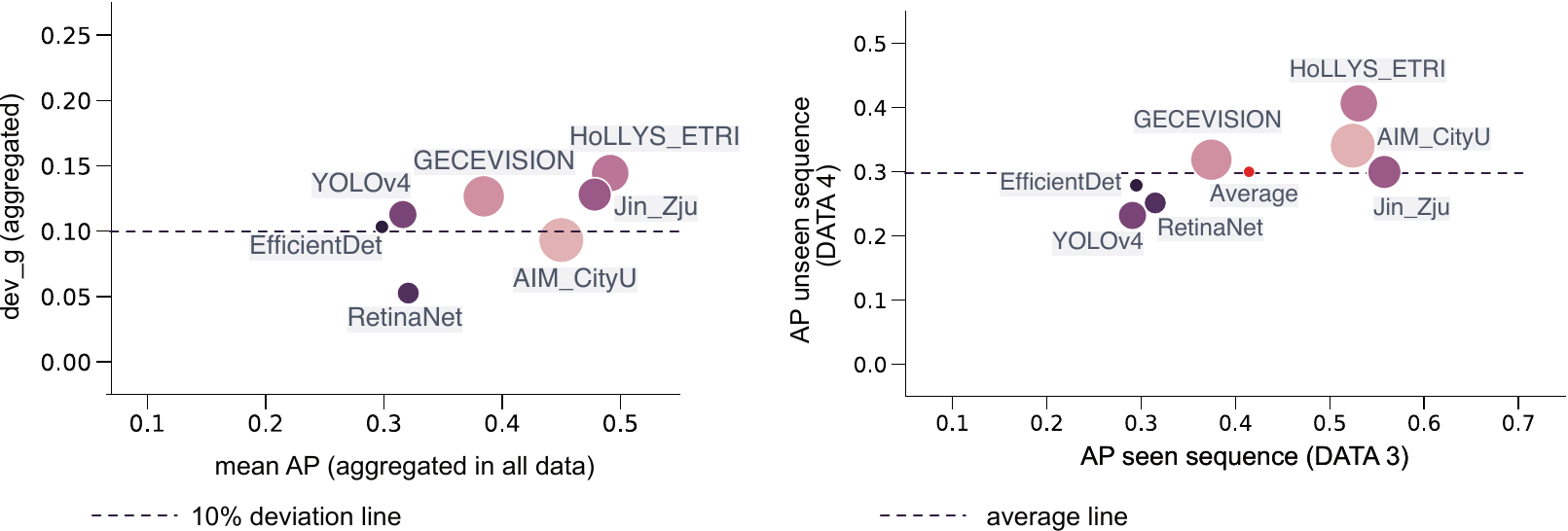}
    \caption{\textbf{Generalisation assessment on detection task:} (left) mean average precision (mAP) on all data versus deviation computed between seen center with unseen modality and unseen center. Least deviation with larger mAP is desired. (right) Comparison of teams and baseline methods on seen center sequence data versus unseen center sequence data, C6. Higher values along both axes is desired.}
    \label{fig:det_my_label}
\end{figure}

\begin{table}[t!]
\small
\centering
\caption{{{Ranking of detection task of polyp generalisation challenge}}}
\begin{tabular}{llllllll}
\toprule
\multirow{2}{*}{\bf Team/Method} &\multirow{2}{*}{\bf Avg\_det $\uparrow$}     & \multicolumn{3}{c}{\bf Avg. deviation scores $\downarrow$}  & & \bf Time $\downarrow$  & \multirow{2}{*}{\bf Rank} $\downarrow$ \\
 \cline{3-6}
 &  & dev\_g$_{1-3}$ & dev\_g$_{2-3}$ & dev\_g$_{4-3}$ & dev\_g & (in s) &  \\\hline
AIM\_CityU~\cite{lia2021joint}     & 0.4501      & 0.0895    & \textbf{0.1339}    & \textbf{0.0562}    & \textbf{0.0932}       &  \textbf{0.10} & \textbf{1}     \\
GECE\_VISION~\cite{polat2021polyp}         & 0.3845      & \textbf{0.0563}    & 0.2537    & 0.0699    & 0.1266        & 0.32  & 5     \\
HoLLYS\_ETRI \cite{honga2021deep}            & \textbf{0.4914}      & 0.1227   & 0.2116    & 0.0988   &0.1444        & 0.69  & \textbf{2}     \\
JIN\_ZJU~\cite{gana2021detection}             & \textbf{0.4783}     & 0.0623    &0.23      & 0.0918   & 0.128       & 1.9 & 3     \\
\hline
YOLOv4~\cite{YOLOv4-Alex2020}           & 0.3161      &0.0998   & 0.1781    & 0.0602    & 0.1127        & 0.13   & 6     \\
RetinaNet (ResNet50) \cite{lin2017focal}     & 0.3207      & \textbf{0.0313}    & \textbf{0.0861}    & \textbf{0.0405}    & \textbf{0.0526}        & 0.27 &  4     \\
EfficientDetD2     & 0.2984      & 0.0586    & 0.1731    & 0.0782    & 0.1033        & 0.20  & 7  \\
 \bottomrule
 \multicolumn{8}{l}{\hspace{.1cm}{$\uparrow$: best increasing} \hspace{.05cm} {$\downarrow$: best decreasing}}
\end{tabular}
\label{detection_table_2}
\end{table}
\subsection*{Aggregated performance and ranking on segmentation task}
\begin{figure}[t!h!]
    \centering
    \includegraphics{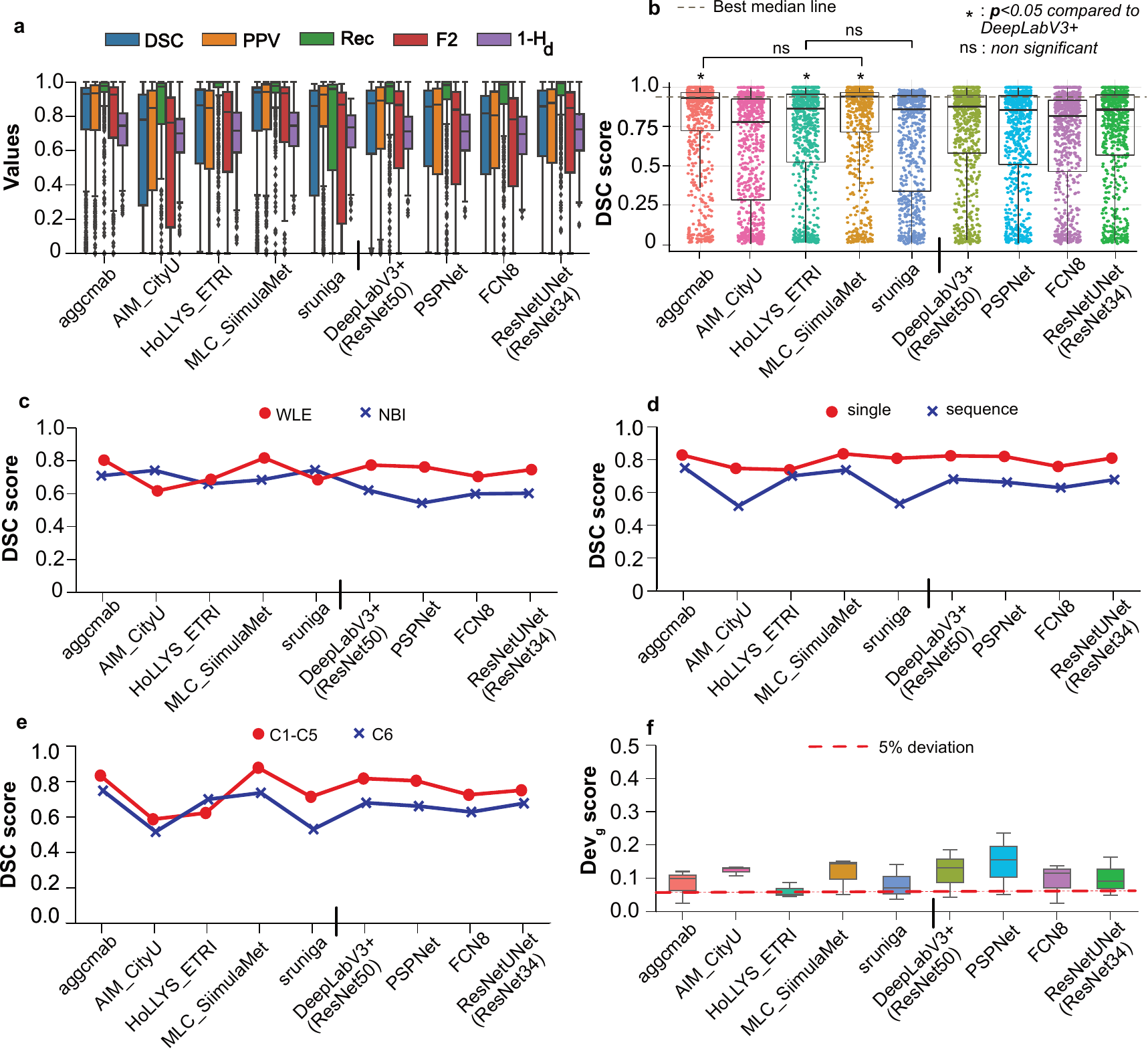}
    \caption{\textbf{Generalisation assessment on segmentation task:} a) Box plots for all segmentation metrics (dice coefficient, DSC; precision, PPV; recall, Rec; F2, type-II error; and Hausdorff distance, Hd) used in the challenge for all test data samples. b) Boxplots representing descriptive statistics over all cases (median, quartiles and outliers) are combined with horizontally jittered dots representing individual data points in all test data. A red-line represent the best median line. It can be observed that teams aggcmab and MLC\_SimulaMet have similar results and with Friedman-Nemenyi post-hoc~\textit{p} value $< 0.05$ denoting significant difference with the best performing baseline DeepLabv3+ method. c) White light endoscopy, WLE versus narrow-band imaging, NBI d) single versus sequence data , e) seen centers, C1-C5 versus unseen center, C6 and f) deviation scores.}
    \label{fig:dsc_all_data_teams_baseline}
\end{figure}

Figure~\ref{fig:dsc_all_data_teams_baseline} (a) demonstrate the boxplots for each teams and baseline methods. It can be observed that the median values for all area-based metrics (dice, precision, recall and F2) are above 0.8 for most teams when compared on all 777 test samples. However, a greater variability can be observed for all teams and baselines that is represented by large number of outlier samples. For the mean distance-based normalised metric ($1-H_d$), only marginal change can be seen for which top teams have higher values as expected. On observing closely only the dice similarity metric in Figure~\ref{fig:dsc_all_data_teams_baseline} (a) where dot and box plots are provided, teams \textit{MLC\_SimulaMet} and \textit{aggcmab} obtained the best scores demonstrating least deviation and with most samples concentrated in the interquartile range (IQR). It can be observed that paired aggcmab and MLC\_SimulaMet; DeepLabV3+(ResNet50) and ResNetUNet(ResNet34); and HoLLYS\_ETRI and PSPNet have similar performances since their quartiles Q1, Q2, and Q3 scores are very close to each other. Although the mean DSC score of team \textit{aggcmab} is slightly higher than the MLC\_SimulaMet, there was no observed statistically significant difference between these two teams. However, both of these teams reported significant difference with $p<0.05$ when compared to the best performing baseline DeepLabV3+(ResNet50).

Tables \ref{table:segmentation_table_1} present the JC, DSC, F2, PPV, Recall, Accuracy and HDF acquired by top five participanting teams and baseline methods (i.e. FCN8, PSPNet, DeepLabV3+ and ResNetUNet) using data 1 to data 4 respectively. As shown in the table for  data 1 (NBI still images), the method suggested by teams \textit{sruniga} and \textit{AIM\_CityU} outperformed against the other teams an baseline methods in terms of JC (>65\%), DSC (> 74\%) and F2 (> 73\%). The team \textit{sruniga} had an outstanding performance in segmenting fewer false-positive regions achieving a PPV result of 81.52 \%  which is higher than other methods by atleast 5\%. Nevertheless, the top recall value for team MLC\_SimulaMet and HoLLYS\_ETRI (> 86\%) proving their ability in detecting more true positive regions. The accuracy results on this data were comparable between all teams and baseline methods ranging from 95.78\% to 97.11\% with the best performance by team \textit{AIM\_CityU}. The results on Data 2 (WLE still images) are also presented in Table \ref{table:segmentation_table_1}. For this data, the methods developed by teams \textit{ MLC\_SimulaMet} and \textit{aggcmab} produced the top values for  JC (>0.77), DSC (>0.82) and F2 (>0.81) with comparable results between two teams. The PPV value was maintained with the method proposed by team \textit{sruniga}(i.e. as discussed for data 1 ) with value of 0.8698 $\pm$ 0.21 followed by team \textit{MLC\_SimulaMet} in second place with a value of 0.8635 $\pm$ 0.26. Additionally, the method by team \textit{MLC\_SimulaMet} surpassed the results for all evaluation measures when compared to the other teams and baseline methods on data 3 as shown in the table. Moreover, the method proposed by team \textit{aggcmab} comes in second place with more the 5\% reduction of results for the JC, DSC and HDF.  For this dataset, the baseline method \textit{DeepLabV3+ (ResNet50)} showed improved performance compared to results on previously discussed data (i.e. data 1 and data 2) where it acquires second place for the F2 and accuracy with a result of 82.66\%  and 95.99\% respectively. Similarly to the performance of the teams on data 3, as shown in Table \ref{table:segmentation_table_1} (i.e. on Data 4 (unseen sequence)) methods by teams \textit{MLC\_SimulaMet} and \textit{aggcmab} produce the best results for most of the evaluation measures JC (>0.68), DSC (>0.73), F2(>0.71), ACC (>0.97) and HDF (<0.34). 
Generally, throughout the evaluation process for all tables on the different datasets, team \textit{sruniga} provided a high PPV value on data 1, data 2 and data 4. Furthermore,  the baseline methods showed low performance in terms of final score compared to the methods proposed by the participants especially with data 1, data 2 and data 4.

To understand the behaviour of each method for provided test data splits we plotted DSC values each separately and compared the ability of methods to generalise on these. From Figure~\ref{fig:dsc_all_data_teams_baseline} (c-d) it can be observed that difference in data setting affect almost all methods. It can be observed that there is nearly upto 20\% gap in performance of the same methods when tested on WLE and NBI. Similarly, for single and sequence frame case and unseen center data. However, it could be observed that those methods that had very close values (e.g., HoLLYS\_ETRI) suffered in performance compared to other methods.

To assess generalisability of each method, we also computed deviation scores for semantic segmentation referred to as $dev\_g$ (see Table~\ref{table:challenge_summary_ranking} and Figure~\ref{fig:dsc_all_data_teams_baseline} (f)). For this assessment, team \textit{aggcmab} ranked the first on both average segmentation scores R$_{seg}$ and deviation score R$_{dev}$. Even though team \textit{ sruniga} was only third on R$_{seg}$, they were second on R$_{dev}$ and ranked at the 1st position for their computation time with average inference time of only 17 ms per second. Team \textit{MLC\_SimulaMet} only was ranked third due to their large computational time of 120 ms per frame and larger deviations (lower generalisation ability). We provide the results of teams with performance below baseline and poor ranking compared to top five teams analysed in the paper in the \textbf{Supplementary Table}~\ref{table:supplmentry_table_1} for completeness. It is to be noted that these teams were selected in the round 3 of the challenge as well but have not been analysed in this paper due to their below baseline scores. 


\nopagebreak
\begin{landscape}
\begin{table}[t!]
\centering
\caption{Team results for the polyp segmentation methods proposed by the participating teams as well as for the baseline methods. All results are  given for data 1, data 2, data 3 and data 4. Top evaluation criteria are highlighted in bold.}
\small
\begin{tabular}{l|l|lllllll}
\toprule
\begin{tabular}[l]{@{}c@{}}\bf{Data}\\ \bf{type} \end{tabular} & \bf{Teams/Method}  &\bf JC $\uparrow$ & \bf DSC $\uparrow$ &\bf F2  $\uparrow$&\bf PPV $\uparrow$& \bf Recall $\uparrow$& \bf ACC $\uparrow$& \bf H$_d$ $\downarrow$ \\ \hline \midrule 
\multirow{9}{*}{\rotatebox[origin=c]{90}{\textbf{{{\begin{tabular}[l]{@{}c@{}}Data 1, NBI-single  \end{tabular}}}}}}

&aggcmab \cite{ galdrana2021multi}	&	 0.6344	$\pm$ {\scriptsize	0.33} 	&	 0.7090	$\pm$ {\scriptsize	0.33	} 	&	 0.7191	$\pm$ {\scriptsize	0.34	} 	&	 0.7521	$\pm$ {\scriptsize	0.34	} 	&	 0.8044	$\pm$ {\scriptsize	0.27	} 	&	 0.9672	$\pm$ {\scriptsize	0.05	} 	&	 0.4411	$\pm$ {\scriptsize	0.20} 	\\

&AIM\_CityU~\cite{lia2021joint}	&	 \textbf{0.6516	$\pm${\scriptsize	0.28}	} 	&\textbf{	 0.7412	$\pm${\scriptsize	0.28	}} 	&	 \textbf{0.7334	$\pm${\scriptsize	0.29	} }	&	\textbf{ 0.7573	$\pm${\scriptsize	0.28	}} 	&	 0.8169	$\pm${\scriptsize	0.26	} 	&	\textbf{ 0.9711	$\pm${\scriptsize	0.05	}} 	&	 0.4067	$\pm${\scriptsize	0.16	} 	\\

&HoLLYS\_ETRI \cite{honga2021deep}	&	 0.5860	$\pm$ {\scriptsize	0.35	} 	&	 0.6584	$\pm$ {\scriptsize	0.36	} 	&	 0.6558	$\pm${\scriptsize	0.36	} 	&	 0.6821	$\pm$ {\scriptsize	0.37	} 	&	\textbf{ 0.8625	$\pm${\scriptsize	0.24	} }	&	 0.9628	$\pm$ {\scriptsize	0.06	} 	&	 0.4350	$\pm$ {\scriptsize	0.21	}
	\\
&MLC\_SimulaMet~\cite{thambawita2021divergentnets}	&	 0.6163	$ \pm${\scriptsize	0.35	} 	&	 0.6839	$\pm$ {\scriptsize	0.36	} 	&	 0.6913	$\pm$ {\scriptsize	0.37	} 	&	 0.7171	$\pm$ {\scriptsize	0.37	} 	&	 \textbf{0.8722	$\pm$ {\scriptsize	0.21	} }	&	 \textbf{0.9677	$\pm$ {\scriptsize	0.06	}} 	& 	 0.4374	$\pm$ {\scriptsize	0.21}	 	\\
&sruniga \cite{ghimirea2021augmentation}	&	 \textbf{0.6672	$\pm$ {\scriptsize	0.31}	} 	&	\textbf{ 0.7441	$\pm$ {\scriptsize	0.31	}} 	&	\textbf{ 0.7507	$\pm$ {\scriptsize	0.31} }	&	 \textbf{0.8152	$\pm$ {\scriptsize	0.27	} }	&	 0.7768	$\pm$ {\scriptsize	0.30	} 	&	 0.9653	$\pm$ {\scriptsize	0.08	} 	&	\textbf{ 0.3711	$\pm$ {\scriptsize	0.16	} }	\\
\cline{2-9}
&{{\begin{tabular}[l]{@{}c@{}}DeepLabV3+ (ResNet50) \cite{chen2017deeplab} \end{tabular}}} 	&	 0.5315	$\pm$ {\scriptsize	0.33	} 	&	 0.6207	$\pm$ {\scriptsize	0.34	} 	&	 0.6005	$\pm$ {\scriptsize	0.36	} 	&	 0.6245	$\pm$ {\scriptsize	0.36	} 	&	 0.8318	$\pm$ {\scriptsize	0.27	} 	&	 0.9642	$\pm$ {\scriptsize	0.06	} 	&	 \textbf{0.3979	$\pm$ {\scriptsize	0.17}	} 	\\

&PSPNet \cite{zhao2017pyramid} 	&	 0.4520	$\pm$ {\scriptsize	0.33	} 	&	 0.5430	$\pm$ {\scriptsize	0.35	} 	&	 0.5189	$\pm$ {\scriptsize	0.36	} 	&	 0.5438	$\pm$ {\scriptsize	0.36	} 	&	 0.7668	$\pm$ {\scriptsize	0.33	} 	&	 0.9559	$\pm$ {\scriptsize	0.05	} 	&	 0.4176	$\pm$ {\scriptsize	0.16	} 	\\
&FCN8 \cite{long2015fully}	&	 0.5054	$\pm$ {\scriptsize	0.32	} 	&	 0.5990	$\pm$ {\scriptsize	0.34	} 	&	 0.6026	$\pm$ {\scriptsize	0.34	} 	&	 0.6440	$\pm$ {\scriptsize	0.35	} 	&	 0.6440	$\pm$ {\scriptsize	0.35	} 	&	 0.9578	$\pm$ {\scriptsize	0.06	} 	&	 0.4330	$\pm$ {\scriptsize	0.17	} 	\\

&{{\begin{tabular}[l]{@{}c@{}}ResNetUNet (ResNet34) \cite{zhang2018road} \end{tabular}}} &	 0.5216	$\pm$ {\scriptsize	0.35	} 	&	 0.6023	$\pm$ {\scriptsize	0.36	} 	&	 0.5839	$\pm$ {\scriptsize	0.38	} 	&	 0.5935	$\pm$ {\scriptsize	0.38	} 	&	 0.8554	$\pm$ {\scriptsize	0.25	} 	&	 0.9634	$\pm$ {\scriptsize	0.05	} 	&	 0.3862$\pm$ {\scriptsize	0.17	}
\\
\bottomrule 

\multirow{9}{*}{\rotatebox[origin=c]{90}{\textbf{{{\begin{tabular}[l]{@{}c@{}}Data 2, WLE-single \end{tabular}}}}}} 
&aggcmab \cite{ galdrana2021multi}	&	 \textbf{0.7701	$\pm$ {\scriptsize	0.27	} }	&	\textbf{ 0.8276	$\pm$ {\scriptsize	0.27	} }	&	\textbf{ 0.8196	$\pm$ {\scriptsize	0.28}	} 	&	 0.8280	$\pm$ {\scriptsize	0.27	} 	&	\textbf{ 0.9231	$\pm$ {\scriptsize	0.16	}} 	&	 0.9829	$\pm$ {\scriptsize	0.04	} 	&	 0.3468	$\pm$ {\scriptsize	0.18	} 	\\
&AIM\_CityU~\cite{lia2021joint}	&	 0.6722	$\pm$ {\scriptsize	0.31	} 	&	 0.7464	$\pm$ {\scriptsize	0.31	} 	&	 0.7391	$\pm$ {\scriptsize	0.31	} 	&	 0.7757	$\pm$ {\scriptsize	0.29	} 	&	 0.8498	$\pm$ {\scriptsize	0.26	} 	&	 0.9647	$\pm$ {\scriptsize	0.11	} 	&	 0.3124	$\pm$ {\scriptsize	0.16	} 	\\
&HoLLYS\_ETRI \cite{honga2021deep}	&	 0.6704	$\pm$ {\scriptsize	0.33	} 	&	 0.7377	$\pm$ {\scriptsize	0.33	} 	&	 0.7229	$\pm$ {\scriptsize	0.33	} 	&	 0.7416	$\pm$ {\scriptsize	0.32	} 	&	 0.9088	$\pm$ {\scriptsize	0.21	} 	&	 0.9709	$\pm$ {\scriptsize	0.08	} 	&	 \textbf{0.3313	$\pm$ {\scriptsize	0.17}	} 	\\
&MLC\_SimulaMet~\cite{thambawita2021divergentnets}	&	\textbf{ 0.7777	$\pm$ {\scriptsize	0.26	}} 	&	\textbf{ 0.8351	$\pm$ {\scriptsize	0.27	} }	&	 \textbf{0.8430	$\pm$ {\scriptsize	0.27	}} 	&	\textbf{ 0.8635	$\pm$ {\scriptsize	0.26	}} 	&	 0.8928	$\pm$ {\scriptsize	0.17	} 	&	 \textbf{0.9852	$\pm$ {\scriptsize	0.02}	} 	&	 \textbf{0.3978	$\pm$ {\scriptsize	0.19	}}	\\
&sruniga \cite{ghimirea2021augmentation}	&	 0.7441	$\pm$ {\scriptsize	0.28	} 	&	 0.8076	$\pm$ {\scriptsize	0.28	} 	&	 0.8066	$\pm$ {\scriptsize	0.28	} 	&	 \textbf{0.8698	$\pm$ {\scriptsize	0.21	}} 	&	 0.8373	$\pm$ {\scriptsize	0.28	} 	&	\textbf{ 0.9841	$\pm$ {\scriptsize	0.02	} }	&	 0.3534	$\pm$ {\scriptsize	0.16	} 	\\

\cline{2-9}
&{{\begin{tabular}[l]{@{}c@{}}DeepLabV3+ (ResNet50) \cite{chen2017deeplab} \end{tabular}}} 	&	 0.7537	$\pm${\scriptsize	0.26	} 	&	 0.8230	$\pm${\scriptsize	0.25	} 	&	 0.8120	$\pm${\scriptsize	0.26	} 	&	 0.8089	$\pm${\scriptsize	0.27	} 	&	 0.9108	$\pm${\scriptsize	0.17	} 	&	 0.9789	$\pm${\scriptsize	0.06	} 	&	 0.3624	$\pm${\scriptsize	0.18	} 	\\
&PSPNet \cite{zhao2017pyramid} 	&	 0.7446	$\pm${\scriptsize	0.25	} 	&	 0.8191	$\pm${\scriptsize	0.24	} 	&	 0.8057	$\pm${\scriptsize	0.25	} 	&	 0.8006	$\pm${\scriptsize	0.25	} 	&	 0.9049	$\pm${\scriptsize	0.17	} 	&	 0.9768	$\pm${\scriptsize	0.06	} 	&	 0.3665	$\pm${\scriptsize	0.19	} 	\\
&FCN8 \cite{long2015fully}	&	 0.6762	$\pm${\scriptsize	0.29	} 	&	 0.7578	$\pm${\scriptsize	0.28	} 	&	 0.7462	$\pm${\scriptsize	0.29	} 	&	 0.7452	$\pm${\scriptsize	0.30	} 	&	 0.9022	$\pm${\scriptsize	0.16	} 	&	 0.9728	$\pm${\scriptsize	0.06	} 	&	 0.4253	$\pm${\scriptsize	0.20	} 	\\

&{{\begin{tabular}[l]{@{}c@{}}ResNetUNet (ResNet34) \cite{zhang2018road} \end{tabular}}}	&	 0.7382	$\pm${\scriptsize	0.27	} 	&	 0.8088	$\pm${\scriptsize	0.26	} 	&	 0.7900	$\pm${\scriptsize	0.27	} 	&	 0.7821	$\pm${\scriptsize	0.28	} 	&	\textbf{ 0.9137	$\pm${\scriptsize	0.20	}} 	&	 0.9767	$\pm${\scriptsize	0.07	} 	&	 0.3295	$\pm${\scriptsize	0.18	} 	\\ 
\bottomrule 

\multirow{9}{*}{\rotatebox[origin=c]{90}{\textbf{{{\begin{tabular}[l]{@{}c@{}}Data 3, seen seq. \end{tabular}}}}}} 
&aggcmab \cite{ galdrana2021multi}	&	\textbf{ 0.7806	$\pm${\scriptsize	0.27	}} 	&	 \textbf{0.8341	$\pm${\scriptsize	0.28	}} 	&	 0.8249	$\pm${\scriptsize	0.28	} 	&	 0.8213	$\pm${\scriptsize	0.29	} 	&	 \textbf{0.9549	$\pm${\scriptsize	0.07}	} 	& 0.9587	$\pm${\scriptsize	0.06	} 	&	 \textbf{0.4517	$\pm${\scriptsize	0.24	}} 	\\

&AIM\_CityU~\cite{lia2021joint}	&	 0.5059	$\pm${\scriptsize	0.36	} 	&	 0.5874	$\pm${\scriptsize	0.36	} 	&	 0.5430	$\pm${\scriptsize	0.37	} 	&	 0.5463	$\pm${\scriptsize	0.36	} 	&	 0.8770	$\pm${\scriptsize	0.29	} 	&	 0.8813	$\pm${\scriptsize	0.13	} 	&	 0.4866	$\pm${\scriptsize	0.25	} 	\\

&HoLLYS\_ETRI \cite{honga2021deep}	&	 0.5433	$\pm${\scriptsize	0.36	} 	&	 0.6229	$\pm${\scriptsize	0.36	} 	&	 0.5953	$\pm${\scriptsize	0.36	} 	&	 0.6072	$\pm${\scriptsize	0.36	} 	&	 0.9081	$\pm${\scriptsize	0.23	} 	&	 0.8906	$\pm${\scriptsize	0.12	} 	&	 0.4804	$\pm${\scriptsize	0.26}	\\ 
&MLC\_SimulaMet~\cite{thambawita2021divergentnets}	&	 \textbf{0.8302	$\pm${\scriptsize	0.23}} 	&	 \textbf{0.8784	$\pm${\scriptsize	0.23	} }	&	\textbf{ 0.8665	$\pm${\scriptsize	0.24	} }	&	 \textbf{0.8602	$\pm${\scriptsize	0.24	}} 	&	\textbf{ 0.9661	$\pm${\scriptsize	0.05	}}	&	 \textbf{0.9752	$\pm${\scriptsize	0.03}} 	&	 \textbf{0.4291	$\pm${\scriptsize	0.22	}} 	\\
& sruniga \cite{ghimirea2021augmentation}	&	 0.6564	$\pm${\scriptsize	0.36	} 	&	 0.7143	$\pm${\scriptsize	0.37	} 	&	 0.7129	$\pm${\scriptsize	0.37	} 	&	 0.7889	$\pm${\scriptsize	0.32	} 	&	 0.7827	$\pm${\scriptsize	0.34	} 	&	 0.9429	$\pm${\scriptsize	0.07} 	&	 0.5310	$\pm${\scriptsize	0.23	} 	\\
\cline{2-9}
&{{\begin{tabular}[l]{@{}c@{}}DeepLabV3+ (ResNet50) \cite{chen2017deeplab} \end{tabular}}} 	&	 0.7459	$\pm${\scriptsize	0.26	} 	&	 0.8176	$\pm${\scriptsize	0.24	} 	&	\textbf{ 0.8266	$\pm${\scriptsize	0.24	}} 	&	 0.8507	$\pm${\scriptsize	0.25	} 	&	 0.8772	$\pm${\scriptsize	0.19	} 	&	 \textbf{0.9599	$\pm${\scriptsize	0.03}}	&	 0.4732	$\pm${\scriptsize	0.22	} 	\\
&PSPNet \cite{zhao2017pyramid} 	&	 0.7316	$\pm${\scriptsize	0.27	} 	&	 0.8050	$\pm${\scriptsize	0.26	} 	&	 0.8183	$\pm${\scriptsize	0.25	} 	&	\textbf{ 0.8527	$\pm${\scriptsize	0.23}	} 	&	 0.8520	$\pm${\scriptsize	0.23	} 	&	 0.9578	$\pm${\scriptsize	0.03	} 	&	 0.5009	$\pm${\scriptsize	0.22	} 	\\
&FCN8 \cite{long2015fully}	&	 0.6258	$\pm${\scriptsize	0.27} 	&	 0.7256	$\pm${\scriptsize	0.27	} 	&	 0.7134	$\pm${\scriptsize	0.26	} 	&	 0.7362	$\pm${\scriptsize	0.27	} 	&	 0.8695	$\pm${\scriptsize	0.23	} 	&	 0.9320	$\pm${\scriptsize	0.06	} 	&	 0.5282	$\pm${\scriptsize	0.23} 	\\

&{{\begin{tabular}[l]{@{}c@{}}ResNetUNet (ResNet34) \cite{zhang2018road} \end{tabular}}}	&	 0.6693	$\pm${\scriptsize	0.29	} 	&	 0.7512	$\pm${\scriptsize	0.29	} 	&	 0.7415	$\pm${\scriptsize	0.29	} 	&	 0.7461	$\pm${\scriptsize	0.31	} 	&	 0.9156	$\pm${\scriptsize	0.17	} 	&	 0.9371	$\pm${\scriptsize	0.06	} 	&	0.5296	$\pm${\scriptsize	0.26}
\\
\bottomrule 

\multirow{9}{*}{\rotatebox[origin=c]{90}{\textbf{{{\begin{tabular}[l]{@{}c@{}}Data 4, unseen seq.  \end{tabular}}}}}} 

&aggcmab \cite{ galdrana2021multi}	&	 \textbf{0.6950	$\pm$ {\scriptsize	0.35	}} 	&	 \textbf{0.7486	$\pm$ {\scriptsize	0.35	}} 	&	\textbf{ 0.7288	$\pm$ {\scriptsize	0.35}	} 	&	 \textbf{0.7542	$\pm$ {\scriptsize	0.34	}} 	&	 \textbf{0.9242	$\pm$ {\scriptsize	0.21	}} 	&	\textbf{ 0.9704	$\pm$ {\scriptsize	0.05	}}	&	\textbf{ 0.3317	$\pm$ {\scriptsize	0.22	}} 	\\
&AIM\_CityU~\cite{lia2021joint}	&	 0.4495	$\pm$ {\scriptsize	0.38	} 	&	 0.5169	$\pm$ {\scriptsize	0.39} 	&	 0.4875	$\pm$ {\scriptsize	0.39	} 	&	 0.6544	$\pm$ {\scriptsize	0.36	} 	&	 0.7011	$\pm$ {\scriptsize	0.42} 	&	 0.9516	$\pm$ {\scriptsize	0.06	} 	&	 0.4347	$\pm$ {\scriptsize	0.19	} 	\\
&HoLLYS\_ETRI \cite{honga2021deep}	&	 0.6370	$\pm$ {\scriptsize	0.36	} 	&	 0.7002	$\pm$ {\scriptsize	0.36	} 	&	 0.6768	$\pm$ {\scriptsize	0.35	} 	&	 0.6930	$\pm$ {\scriptsize	0.35	} 	&	 \textbf{0.9143	$\pm$ {\scriptsize	0.23	}} 	&	 0.9636	$\pm$ {\scriptsize	0.04	} 	&	 0.3913	$\pm$ {\scriptsize	0.26	} 	\\
&MLC\_SimulaMet~\cite{thambawita2021divergentnets}&\textbf{ 0.6835	$\pm$ {\scriptsize	0.36}} 	&	\textbf{ 0.7367	$\pm$ {\scriptsize	0.35	}} 	&	\textbf{ 0.7187	$\pm$ {\scriptsize	0.36	}} 	&	 0.7187	$\pm$ {\scriptsize	0.36} 	&	 0.9099	$\pm$ {\scriptsize	0.23} 	&	 \textbf{0.9719	$\pm$ {\scriptsize	0.05}} 	&	 \textbf{0.3353	$\pm$ {\scriptsize	0.22}} 	\\
& sruniga \cite{ghimirea2021augmentation}	&	 0.4718	$\pm$ {\scriptsize	0.39	} 	&	 0.5316	$\pm$ {\scriptsize	0.41	} 	&	 0.5097	$\pm$ {\scriptsize	0.41	} 	&	 \textbf{0.7528	$\pm$ {\scriptsize	0.32	}} 	&	 0.6489	$\pm$ {\scriptsize	0.44	} 	&	 0.9654	$\pm$ {\scriptsize	0.05	} 	&	 0.4522	$\pm$ {\scriptsize	0.17} 	\\
\cline{2-9}
& {{\begin{tabular}[l]{@{}c@{}}DeepLabV3+ (ResNet50) \cite{chen2017deeplab} \end{tabular}}} 	&	 0.6126	$\pm$ {\scriptsize	0.36	} 	&	 0.6803	$\pm$ {\scriptsize	0.36	} 	&	 0.6575	$\pm$ {\scriptsize	0.36	} 	&	 0.7182	$\pm$ {\scriptsize	0.33	} 	&	 0.8518	$\pm$ {\scriptsize	0.29	} 	&	 0.9637	$\pm$ {\scriptsize	0.05	} 	&	 0.3814	$\pm$ {\scriptsize	0.22	} 	\\
 &PSPNet \cite{zhao2017pyramid} 	&	 0.5973	$\pm$ {\scriptsize	0.37} 	&	 0.6616	$\pm$ {\scriptsize	0.37	} 	&	 0.6316	$\pm$ {\scriptsize	0.38	} 	&	 0.6763	$\pm$ {\scriptsize	0.35	} 	&	 0.8722	$\pm$ {\scriptsize	0.29	} 	&	 0.9622	$\pm$ {\scriptsize	0.05	} 	&	 0.3438	$\pm$ {\scriptsize	0.22	} 	\\
 &FCN8 \cite{long2015fully}	&	 0.5615	$\pm$ {\scriptsize	0.37	} 	&	 0.6285 $\pm$ {\scriptsize	0.38	} 	&	 0.5970	$\pm$ {\scriptsize	0.38	} 	&	 0.6505	$\pm$ {\scriptsize	0.36	} 	&	 0.8623	$\pm$ {\scriptsize	0.30	} 	&	 0.96	$\pm$ {\scriptsize	0.05} 	&	 0.3634	$\pm$ {\scriptsize	0.23	} 	\\

&{{\begin{tabular}[l]{@{}c@{}}ResNetUNet (ResNet34) \cite{zhang2018road} \end{tabular}}}	&	 0.6140	$\pm$ {\scriptsize	0.36	} 	&	 0.6777	$\pm$ {\scriptsize	0.36	} 	&	 0.6516	$\pm$ {\scriptsize	0.37	} 	&	 0.7097	$\pm$ {\scriptsize	0.33	} 	&	 0.8781	$\pm$ {\scriptsize	0.28	} 	&	 0.9653	$\pm$ {\scriptsize	0.04	} 	&	 0.3993	$\pm$ {\scriptsize	0.24	} 	\\
\bottomrule 
 \multicolumn{6}{l}{\hspace{.1cm}{$\uparrow$: best increasing} \hspace{.05cm} {$\downarrow$: best decreasing}}
\end{tabular}
\label{table:segmentation_table_1}
\end{table}
\end{landscape}
\begin{table*} [t!]
\small
\centering
\caption{{\textbf{Ranking of segmentation task of polyp generalisation challenge:} Ranks are provided based on a) semantic score aggregation, R$_{seg}$; b) average deviation score,  R$_{dev}$; and c) overall ranking (R$_{all}$) that takes into account R$_{seg}$, R$_{dev}$ and time. For ties in the final ranking (R$_{all}$), segmentation score is taken into account. For time, ranks are provided into three categories: teams with $< 50$ ms, between $50-100$ ms and $> 100$ ms. Top two values are in bold.}}
\begin{tabular}{l|lllllllccc}
\toprule
\multirow{2}{*}{\bf Team/Method} & \multicolumn{3}{c}{\bf Average Seg\_score $\uparrow$}   & \multicolumn{3}{c}{\bf Average Dev\_score $\downarrow$}  & \bf Time $\downarrow$  & \bf R$_{seg}$ $\downarrow$ & \bf R$_{dev}$  $\downarrow$& \bf R$_{all}$ $\downarrow$\\
 \cline{2-8}
  & \bf Data 1 & \bf  Data 2 &\bf  Data 4  &\bf dev\_g$_{1-3}$& \bf dev\_g$_{2-3}$  & \bf dev\_g$_{4-3}$  &\bf (ms) &\bf (avg.) &\bf (avg.)&\bf (avg.) \\
  \hline
 aggcmab \cite{ galdrana2021multi} & 0.7461 & \textbf{0.8496} & \textbf{0.7889} & 0.1199 & \textbf{0.0244} & 0.0994 & 107 & \textbf{1}  & \bf{1} & \bf{1}\\
 AIM\_CityU~\cite{lia2021joint} & \textbf{0.7621} & 0.7777 & 0.5899 &0.1289 &0.1322 & 0.1072 & 80 & 4  & 3 & 5 \\
 HoLLYS\_ETRI \cite{honga2021deep} & 0.7146 & 0.7777 & 0.746 & \textbf{0.0453} & 0.0864 & \textbf{0.0499}& 84 & 5  & \bf{1} & 4\\
 MLC\_SimulaMet~\cite{thambawita2021divergentnets} &0.7411& \textbf{0.8586} & \textbf{0.7813} & 0.151 & 0.0508 & 0.1425 & 120 & \bf{2} & 3 & 3  \\
 sruniga \cite{ghimirea2021augmentation} & \textbf{0.7716} & 0.8303 & 0.6107 & \textbf{0.0357}& 0.0703 & 0.1411 & \textbf{17} & {3}  & 2 & \bf{2}\\
 \hline
  \multicolumn{9}{l}{\textbf{Baselines}}   \\
  \hline
 DeepLabV3+ (ResNet50) \cite{chen2017deeplab} & 0.6693 & 0.8387 & 0.7269 & 0.1847 & 0.0422 & 0.1305 & 19 & NA  & NA& NA \\
PSPNet \cite{zhao2017pyramid,chen2017deeplab} & 0.5931 & 0.8325 & 0.7104 & 0.2356 & 0.0504 & 0.1558 & 45 & NA  & NA & NA\\
FCN8 \cite{long2015fully} & 0.6508 & 0.7878 & 0.6845 & 0.1376 & \textbf{0.0248} & 0.1152 & 27 &NA  & NA & NA\\
 ResNetUNet-ResNet34 \cite{zhang2018road} & 0.6587 & 0.8236 & 0.7293 & 0.1627 & 0.0485 & \textbf{0.0903} &\textbf{13} & NA  & NA & NA\\
 \bottomrule
 \multicolumn{9}{l}{\hspace{.1cm}{$\uparrow$: best increasing} \hspace{.05cm} {$\downarrow$: best decreasing}}
\end{tabular}
 \label{table:challenge_summary_ranking}
\end{table*}
%
%
\begin{figure}[t!]
    \centering
    \includegraphics[scale=0.65]{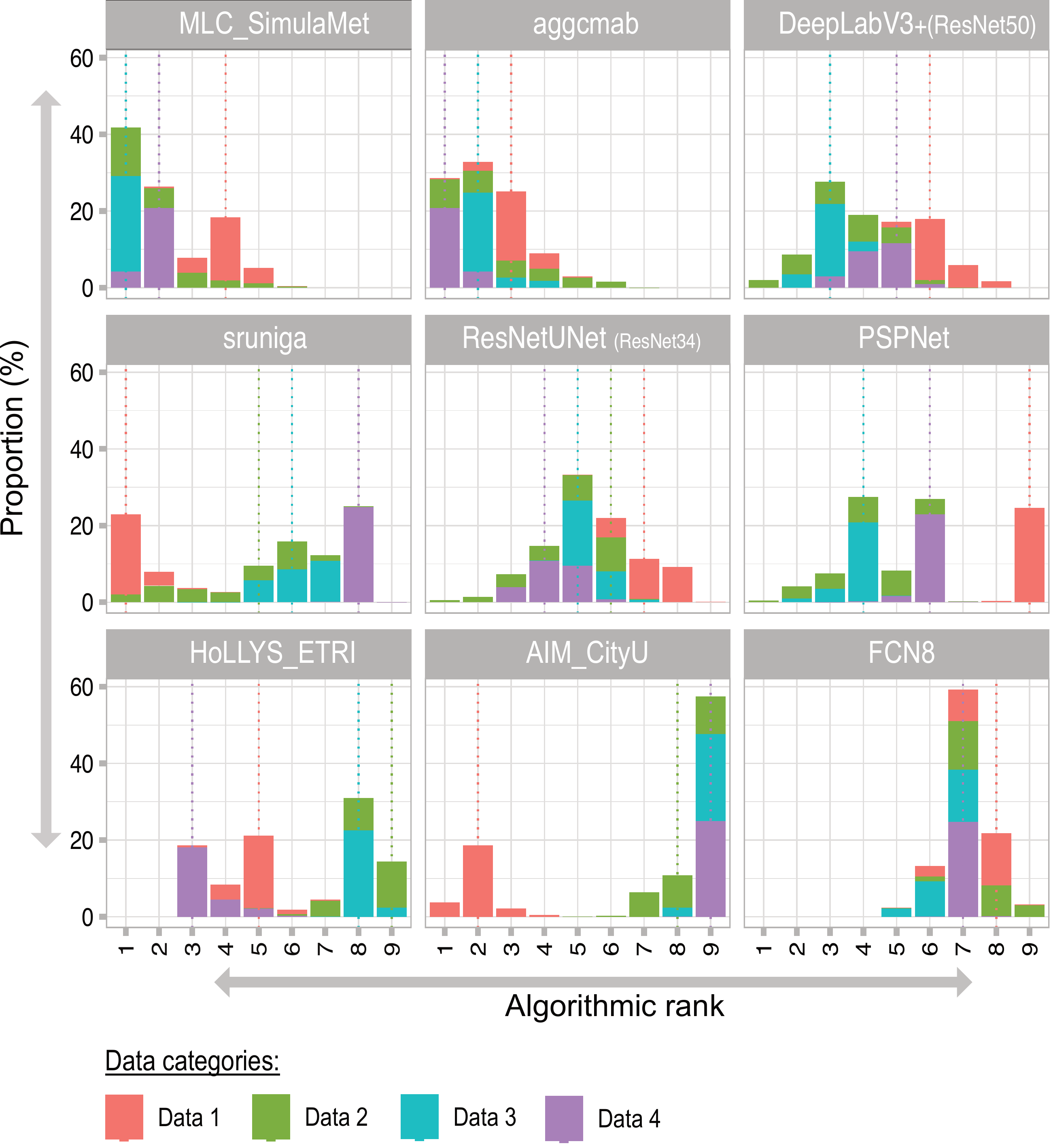}
    \caption{\textbf{Algorithmic rank-based on across bootstrap samples~\cite{Wiesenfarth2021} are displayed with colours according to the data categories:} Histogram bars show how much proportion (in \%) of each data contribute to the ranking of each team and baseline method. Here, for ranking we have only considered dice similarity coefficient values.}
    \label{fig:dsc_proportion}
\end{figure}
\section*{Discussion}
While polyp detection and segmentation using computer vision methods, in particular deep learning, have been widely studied in the past, rigorous assessment and benchmarking on centerwise split, modality split and sequence data have not been comprehensively studied. In our EndoCV2021 edition, we challenged participants to address generalisability issues in polyp detection and segmetation methods by providing multicenter and diverse data. 
For polyp detection and localisation, 3/4 teams chose feature pyramid-based networks while one team used YOLOV5 ensemble paradigm. Unlike most of these methods that require anchors to detect multiple objects of different scales, and overlap, team AIM\_CityU~\cite{lia2021joint} used an anchor free fully convolution one-stage object detection (FCOS) method. HoLLYS\_ETRI \cite{honga2021deep} focused mostly on accuracy and used an ensemble to train five different models, i.e., one model per center, and an aggregated model output was devised for the test inference. Even though, the HoLLYS\_ETRI team led the leaderboard ranking on the average detection score on almost all data type, the observed detection speed (0.69 sec.) and the deviation in generalisation score only put them on the second rank (see Table~\ref{detection_table_2}). On contrary, AIM\_CityU team with their anchor free single stage network performed consistently well in almost all data with the fastest inference (0.1 sec.) and the least deviation score (see Figure~\ref{fig:det_my_label} and Table~\ref{detection_table_2}) between teams, and hence leading the leaderboard.

\textit{Hypothesis I: It can be hypothesised that anchor free detection methods can better generalise compared to methods that require anchors in heterogeneous multi-center dataset. This is strictly true as the polyp sizes in the dataset is varied (see Figure~\ref{fig:my_label_data} b) and also the image sizes ranged from $388 \times 288$ pixels to $1920 \times 1080$.}

%
Since, all methods trained their algorithm on single frame images provided in this challenge, it can be observed in Table~\ref{detection_table_1} that the detection scores for all methods are relatively higher for the data 2 (WLE-single) , compared to other data categories, despite they came from unseen data center 6. However, performance drop can be observed for both seen (centers, C1-C5) and unseen (center, C6) sequence data that consisted of WLE images only. In addition, change in modality has detrimental effect on the performance for all methods even on single frames (see for data 1, NBI-single, Table~\ref{detection_table_1}). 

\textit{Hypothesis II: It can be hypothesised that methods trained on single frame produce sub-optimal and inconsistent detection in videos as image-based object detection cannot leverage the rich temporal information inherent in video data. The scenario gets worsen when applied on different center to that on which it was trained. To overcome this, Long Short-Term Memory (LSTM) based methods can be used to keep the temporal information for pruning predictions~\cite{YongyiCVPR2017:DISCUSSION}. }


For segmentation task, while most teams used ensemble technique targeting to win on the leaderboard (MLC\_SimulaMet~\cite{thambawita2021divergentnets}, HoLLYS\_ETRI \cite{honga2021deep}, aggcmab~\cite{ galdrana2021multi}), there were some teams who worked towards model efficiency network (e.g., team sruniga \cite{ghimirea2021augmentation}) or modifications for faster inference and improved accuracy (e.g., team AIM\_CityU~\cite{lia2021joint}). Light weight model using HarDNet68 backbone with aggregated maps across scales (team sruniga) and use of multi-scale feature fusion network (HRNet) with low-rank disentanglement by team AIM\_CityU outperformed all other methods on narrow band imaging modality (data 1) including the baseline segmentation methods (see Table~\ref{table:segmentation_table_1}). These methods showed acceptable performance for single frames on unseen data (data 2, WLE-single) as well. However, on sequence data (both for seen sequence data 3 and unseen sequence data 4), both of these methods performed poorly compared to ensemble-based techniques (see Figure~\ref{fig:dsc_all_data_teams_baseline} (c-e)). Several network conjoint by MLC\_SimulaMet and dual UNet network used by the team aggcmab have disadvantage of large inference time (nearly 6 times higher than the fastest method). 
%

%
%

\textit{Hypothesis III: It can be hypothesised that on single frame data multi-scale feature fusion networks perform better irrespective of their modality changes. This is without requiring to ensemble same or multiple models for inference which ideally increases both model complexity and inference time. However, on sequence data we advice to incorporate temporal information propagation in the designed networks. Furthermore, to improve model generalisation on unseen modality, domain adaptation techniques can be applied~\cite{Celik2021:MICCAI}.}

HoLLYS\_ETRI \cite{honga2021deep} used instance segmentation approach with five separate models trained on C1 to C5 training data separately. It can be observed that this scheme provided better generalisation ability in most cases leading to the least deviation on average dice score (see Figure~\ref{fig:dsc_all_data_teams_baseline} (f)). However, reported dice metric values were lower than most methods especially ensemble techniques that are targeted towards higher accuracy but are less generalisable (in terms of consistency in test inference across multiple data categories). {This is also evident in Figure~\ref{fig:dsc_proportion} where proportion of samples from data 1 for top performing teams \textit{aggcmab} and \textit{MLC\_SimulaMet} are mostly ranked on the third and fourth ranks.} 


\textit{Hypothesis IV: It can be hypothesised that pretext tasks can lead to improved generalisability. However, to boost model accuracy, modifications are desired that could include feature fusion blocks and other aggregation techniques.}


\section*{Conclusion}
We provided a comprehensive dissection of widely used deep learning baseline methods and methods devised by top participants in crowd-sourcing initiative of EndoCV2021 challenge. Through our experimental design, provided multi-center dataset and holistic comparisons, we demonstrate the need of generalisable methods to tackle real-world clinical challenges required for robust polyp detection and segmentation tasks. While most methods provided improvement over several widely used baseline methods, their design adversely impacted algorithmic robustness and/or real-time capability when provided unseen sequence data and different modality. A better trade-off in both inference time and generalisability is the key take away of this work. We provide experimental-based hypothesis to encourage future research towards innovating more applicable methods that can work effectively in multi-center data and diverse modalities that are widely used in colonoscopic procedures. 

\paragraph*{Author contributions}
S. Ali conceptualised the work, led the challenge and workshop, prepared the dataset, software and performed all analyses. S. Ali, N. Ghatwary and D. Jha contributed in data annotations. T. de Lange, J.E. East, S. Realdon, R. Cannizzaro, D. Lamarque were involved providing colonoscopy data and in the validation and quality checks of the annotations used in this challenge. Challenge participants (E. Isik-Polat, G. Polat, C. Yang, S. Poudel, S. Hicks, Z. Jin, T. Gan, C. Yu, D. Yeo, M. Haitimi) were involved in method summary and compilation of the related work. E. Isik-Polat performed the statistical tests conducted in this paper. S. Ali wrote most of the manuscript with inputs from N. Ghatwary and all co-authors. All authors participated in the revision of this manuscript, provided input and agreed for submission.

\paragraph*{Declaration of Competing Interest}
The authors declare that they have no known competing financial interests or personal relationships that could have appeared to influence the work reported in this paper.

\paragraph*{Acknowlgedgments}
The research was supported by the National Institute for Health Research (NIHR) Oxford Biomedical Research Centre (BRC). The views expressed are those of the authors and not necessarily those of the NHS, the NIHR or the Department of Health. 

\section*{Supplementary Material}
\section*{Supplementary Tables}
\begin{suppTable*}[t!h!]
	\centering
	\begin{tabular}{l|l|l|l|l|l|l|l} 
		\hline
\bf Dataset &\bf Findings & \bf \# of samples & \bf Resolution & \bf Modality & \bf \begin{tabular}[t]{@{}l@{}}Study\\ type\end{tabular} & \bf Challenge & \bf Availability \\ \hline
		CVC-ColonDB~\cite{bernal2012towards}& Polyps & 380 images$^\dag$ $^\dagger$  &$574\times500$ &WLE  &Single & APC & by request$^\bullet$ \\ 
		\hline
				
		CVC-ClinicDB~\cite{bernal2015wm} & Polyps & 612 images$^\dag$& 384$\times$288  &WLE &Single &EndoVis
 & \begin{tabular}[t]{@{}l@{}}open\\academic \end{tabular} \\ \hline 
		CVC-VideoClinicDB~\cite{bernal2017miccai} & Polyps & 11,954 images$^\dag$ & 384$\times$288 & WLE &Single  &EndoVis
 & by request$^\bullet$ \\ \hline
 	    EDD2020~\cite{ali2020endoscopy,ali2021_endoCV2020} & \begin{tabular}[t]{@{}l@{}} GI findings \\ with polyps\end{tabular} & 386 images  & variable& \begin{tabular}[t]{@{}l@{}}WLE,\\ NBI\end{tabular} & Multi & \begin{tabular}[t]{@{}l@{}}EndoCV\\(2020) \end{tabular}& \begin{tabular}[t]{@{}l@{}}open\\ academic\end{tabular}\\ \hline
		ETIS-Larib Polyp DB~\cite{silva2014toward} & Polyps & 196 images$^\dag$  &1224$\times$966 & WLE &Single  & EndoVis
 & \begin{tabular}[t]{@{}l@{}}open\\academic \end{tabular}\\ \hline
		\begin{tabular}[t]{@{}l@{}} ASU-Mayo polyp  \\ database~\cite{tajbakhsh2015automated}\end{tabular}  & Polyps & 18,781 images$^\dag$ &$688\times 550$   & WLE &Single  & EndoVis & by request$^\bullet$  \\ \hline
		HyperKvasir~\cite{borgli2020hyperkvasir} & \begin{tabular}[t]{@{}l@{}} GI findings \\ with polyps\end{tabular} & \begin{tabular}[t]{@{}l@{}} 110,079 images \\ \& 374 videos\end{tabular} &\begin{tabular}[t]{@{}l@{}}$720\times 576$\\to $1920\times 1072$\end{tabular}  & WLE &Single & NA & \begin{tabular}[t]{@{}l@{}}open\\ academic\end{tabular} \\ \hline
		Kvasir-SEG~\cite{jha2020kvasir} & Polyps & 1000 images$^\dag$& \begin{tabular}[t]{@{}l@{}}  $332 \times 487$ \\ , $1920\times1072$\end{tabular} & WLE  & Single &\begin{tabular}[t]{@{}l@{}} Medico\\ MedAI\end{tabular} & \begin{tabular}[t]{@{}l@{}}open\\ academic\end{tabular}  \\ \hline
		PolypGen~\cite{ali2021polypgen} & \begin{tabular}[t]{@{}l@{}}  Polyp \\non-polyp \end{tabular} &  \begin{tabular}[t]{@{}l@{}} 3446 images \\ including \\sequence data$^\dag$\end{tabular} & \begin{tabular}[t]{@{}l@{}} $384\times 288$\\ to $1920 \times 1080$\end{tabular} &\begin{tabular}[t]{@{}l@{}}  WLE, \\NBI\\ (test) \end{tabular} & Multi & \begin{tabular}[t]{@{}l@{}}EndoCV\\(2021)\end{tabular} & \begin{tabular}[t]{@{}l@{}} open\\ academic \end{tabular} \\ \hline
	  \multicolumn{7}{l}{$^\dag$\footnotesize{Including ground truth segmentation masks} \hspace{.1cm}
	   $^\ddagger$\footnotesize{Contour} \hspace{.1cm}
	   $^\diamond$\footnotesize{Video capsule endoscopy}\hspace{.1cm}
	   $^\bullet$\footnotesize{Not available anymore or unknown}}\\
	    \multicolumn{7}{l}{$^\clubsuit$\footnotesize{Medical atlas for education with several low-quality samples of various GI findings} \hspace{.1cm}} 
	    \\
	    \multicolumn{7}{l}{\footnotesize{APC: Automatic polyp classification; EndoVis: MICCAI Endoscopic vision challenge; EndoCV: IEEE ISBI Endoscopic challenge} \hspace{.1cm}}
	 \end{tabular}
     \caption{An overview of existing gastrointestinal lesion datasets including polyps}  
	\label{tab:datasets}
\end{suppTable*}

\begin{suppfigure}
    \centering
        \includegraphics[width=0.95\textwidth]{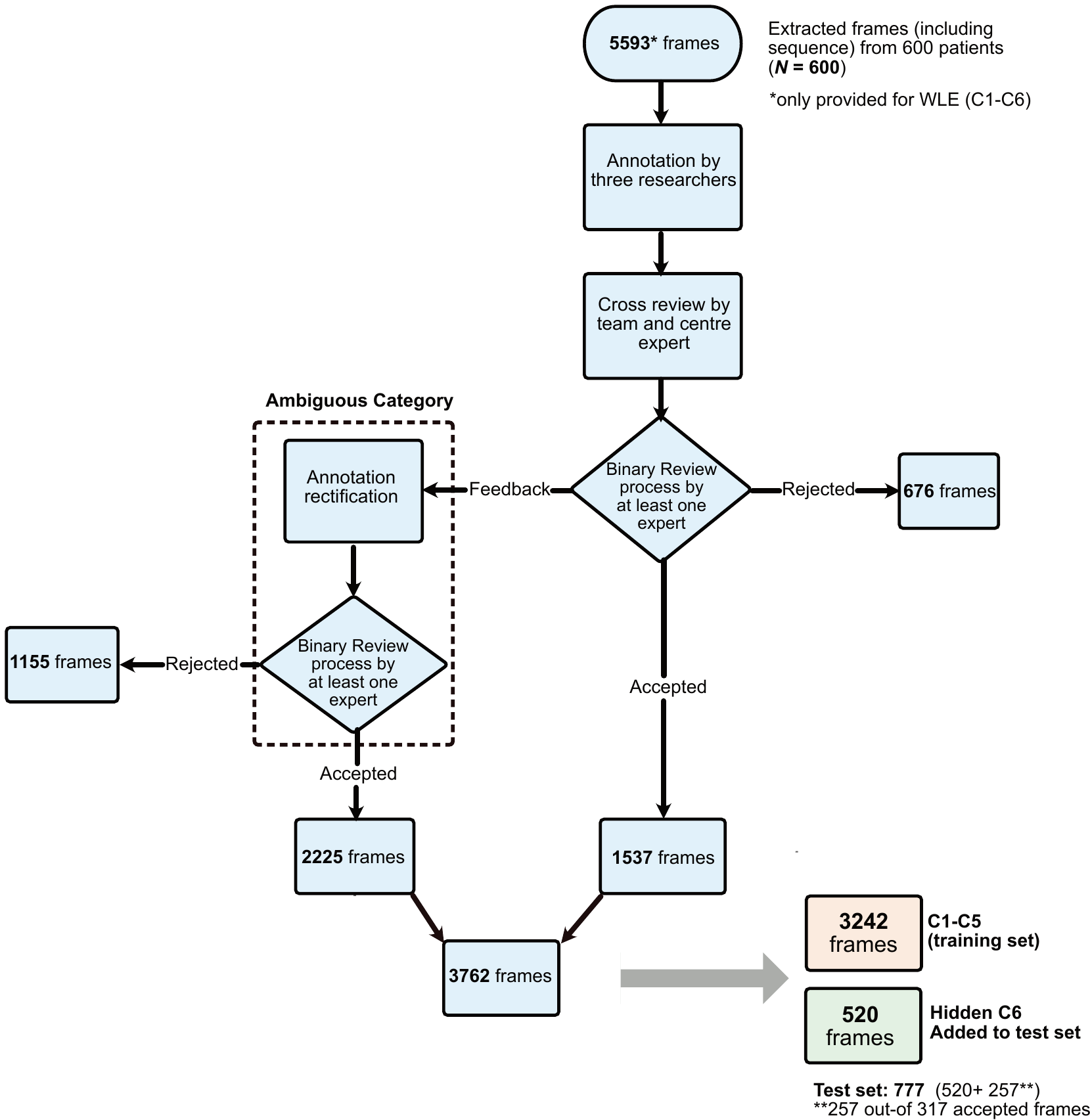}
    \caption{\textbf{Annotation workflow:} 600 patients (N=600) data was used that consisted of both videos and frames. First 5593 relevant frames for polyp detection and segmentation were extracted. These frames comprised of both single and sequence data. For details please see Figure~\ref{fig:my_label_data}. Review of annotations was done by at least one expert and the frames were either relabeled or immediately rejected. A second review was conducted by at least one expert. Here expert refers to a senior consultant gastroenterologist. Overall, 3762/5593 frames were annotated of which 520 frames from center 6 was directly embedded in the test set. For testing set, a similar strategy was taken for which 257 samples out-of 317 samples were accepted during the review phase.}
    \label{fig:suppl_data_annotation}
\end{suppfigure}


\begin{landscape}
\begin{suppTable}[t!]
\centering
\caption{Semantic segmentation results for teams ranking below 5th place on out-of-sample data 1, data 2, data 3 and data 4.}
\begin{tabular}{l|l|lllllll}
\toprule
\begin{tabular}[l]{@{}c@{}}\bf{Data}\\ \bf{type} \end{tabular} & \bf{Teams/Method}  &\bf JC $\uparrow$ & \bf DSC $\uparrow$ &\bf F2  $\uparrow$&\bf PPV $\uparrow$& \bf Recall $\uparrow$& \bf ACC $\uparrow$& \bf H$_d$ $\downarrow$ \\ \hline \midrule 
\multirow{4}{*}{\rotatebox[origin=c]{90}{\textbf{{{\begin{tabular}[l]{@{}c@{}} Data 1 \\ (NBI-single)  \end{tabular}}}}}}
& \\
&YCH\_THU \cite{yua2021parallel}	&	0.2625	$\pm${\scriptsize	0.2917	}	&	0.3403	$\pm${\scriptsize	0.3314	}	&	0.3831	$\pm${\scriptsize	0.3563	}	&	0.5181	$\pm${\scriptsize	0.4149	}	&	0.3426	$\pm${\scriptsize	0.3473	}	&	0.8767	$\pm${\scriptsize	0.0971	}	&	0.5443	$\pm${\scriptsize	0.1831	}	\\
&Mah\_UNM \cite{haithamia2021embedded}	&	03275	$\pm${\scriptsize	0.3121}	&	0.4127	$\pm${\scriptsize	0.3488	}	&	0.4045	$\pm${\scriptsize0.3588	}	&	0.4307	$\pm${\scriptsize	0.3870	}	&	0.6962	$\pm${\scriptsize	0.3496	}	&	0.9463	$\pm${\scriptsize	0.0626	}	&	0.4119	$\pm${\scriptsize	0.1501	}	\\
&NDS\_MultiUni\cite{tomara2021improving}	&	0.1765	$\pm${\scriptsize	0.2569	}	&	0.2371	$\pm${\scriptsize	0.2944	}	&	0.2598	$\pm${\scriptsize	0.3145	}	&	0.3165	$\pm${\scriptsize	0.3855	}	&	0.5905	$\pm${\scriptsize	0.4158	}	&	0.9116	$\pm${\scriptsize	0.0740	}	&	0.4968	$\pm${\scriptsize	0.1758	}	\\	\\
\bottomrule 

\multirow{4}{*}{\rotatebox[origin=c]{90}{\textbf{{{\begin{tabular}[l]{@{}c@{}} Data 2 \\ (WLE-single)  \end{tabular}}}}}}
& \\
& YCH\_THU \cite{yua2021parallel}	&	0.5136	$\pm${\scriptsize	0.3444	}	&	0.5986	$\pm${\scriptsize	0.3566	}	&	0.6402	$\pm${\scriptsize	0.3568	}	&	0.7668	$\pm${\scriptsize	0.3265	}	&	0.5749	$\pm${\scriptsize	0.3707	}	&	0.9337	$\pm${\scriptsize	0.0845	}	&	0.4868	$\pm${\scriptsize	0.2035	}\\
& Mah\_UNM \cite{haithamia2021embedded}	&	0.4726	$\pm${\scriptsize	0.3166	}	&	0.5698	$\pm${\scriptsize	0.3403	}	&	0.5885	$\pm${\scriptsize	0.3467	}	&	0.6425	$\pm${\scriptsize	0.3575	}	&	0.6523	$\pm${\scriptsize	0.3424	}	&	0.9471	$\pm${\scriptsize	0.0770	}	&	0.4599	$\pm${\scriptsize	0.1805	}\\
& NDS\_MultiUni\cite{tomara2021improving}	&	0.3373	$\pm${\scriptsize	0.2784	}	&	0.4404	$\pm${\scriptsize	0.3134	}	&	0.4593	$\pm${\scriptsize	0.3200	}	&	0.5426	$\pm${\scriptsize	0.3643	}	&	0.5480	$\pm${\scriptsize	0.3764	}	&	0.9182	$\pm${\scriptsize	0.0954	}	&	0.5533	$\pm${\scriptsize	0.2026	}
\\ \\
\bottomrule 

\multirow{4}{*}{\rotatebox[origin=c]{90}{\textbf{{{\begin{tabular}[l]{@{}c@{}} Data 3 \\ (seen seq.)  \end{tabular}}}}}}
& \\
& YCH\_THU \cite{yua2021parallel}	&	0.4994	$\pm${\scriptsize	0.3158	}	&	0.5988	$\pm${\scriptsize	0.3252	}	&	0.6494	$\pm${\scriptsize	0.3359	}	&	0.7958	$\pm${\scriptsize	0.3197	}	&	0.5857	$\pm${\scriptsize	0.3535	}	&	0.8925	$\pm${\scriptsize	0.0909	}	&	0.5926	$\pm${\scriptsize	0.1622	}	\\
& Mah\_UNM \cite{haithamia2021embedded}	&	0.4271	$\pm${\scriptsize	0.3491	}	&	0.5092	$\pm${\scriptsize	0.3712	}	&	0.5364	$\pm${\scriptsize	0.3978	}	&	0.5895	$\pm${\scriptsize	0.4320	}	&	0.7131	$\pm${\scriptsize	0.3317	}	&	0.8652	$\pm${\scriptsize	0.1329	}	&	0.5488	$\pm${\scriptsize	0.2409	}	\\
& NDS\_MultiUni\cite{tomara2021improving}	&	0.5264	$\pm${\scriptsize	0.3234	}	&	0.6240	$\pm${\scriptsize	0.3150	}	&	0.6744	$\pm${\scriptsize	0.3011	}	&	0.7920	$\pm${\scriptsize	0.2973	}	&	0.6509	$\pm${\scriptsize	0.3300	}	&	0.9149	$\pm${\scriptsize	0.0612	}	&	0.6708	$\pm${\scriptsize	0.1963	}	\\
\\
\bottomrule 

\multirow{4}{*}{\rotatebox[origin=c]{90}{\textbf{{{\begin{tabular}[l]{@{}c@{}}Data 4 \\ (unseen seq.)  \end{tabular}}}}}}
& \\
& YCH\_THU \cite{yua2021parallel}	&	0.3285	$\pm${\scriptsize	0.3204	}	&	0.4096	$\pm${\scriptsize	0.3577	}	&	0.4443	$\pm${\scriptsize	0.3820	}	&	0.7475	$\pm${\scriptsize	0.3417	}	&	0.4143	$\pm${\scriptsize	0.3720	}	&	0.8802	$\pm${\scriptsize	0.1179	}	&	0.8802	$\pm${\scriptsize	0.1179	}	\\
& Mah\_UNM \cite{haithamia2021embedded}	&	0.2491	$\pm${\scriptsize	0.3219	}	&	0.3064	$\pm${\scriptsize	0.3635	}	&	0.3023	$\pm${\scriptsize	0.3715	}	&	0.5304	$\pm${\scriptsize	0.4321	}	&	0.4596	$\pm${\scriptsize	0.4260	}	&	0.9281	$\pm${\scriptsize	0.0742	}	&	0.5244	$\pm${\scriptsize	0.1970	}	\\
& NDS\_MultiUni\cite{tomara2021improving}	&	0.2491	$\pm${\scriptsize	0.3386	}	&	0.3046	$\pm${\scriptsize	0.3537	}	&	0.2941	$\pm${\scriptsize	0.3543	}	&	0.4297	$\pm${\scriptsize	0.4139	}	&	0.5290	$\pm${\scriptsize	0.4182	}	&	0.9202	$\pm${\scriptsize	0.0749	}	&	0.9202	$\pm${\scriptsize	0.0749	}	
\\
	\\
\bottomrule 
 \multicolumn{6}{l}{\hspace{.1cm}{$\uparrow$: best increasing} \hspace{.05cm} {$\downarrow$: best decreasing}}
\end{tabular}
\label{table:supplmentry_table_1}
\end{suppTable}
%
\end{landscape}
\bibliography{sample}

\begin{thebibliography}{10}
\urlstyle{rm}
\expandafter\ifx\csname url\endcsname\relax
  \def\url#1{\texttt{#1}}\fi
\expandafter\ifx\csname urlprefix\endcsname\relax\def\urlprefix{URL }\fi
\expandafter\ifx\csname doiprefix\endcsname\relax\def\doiprefix{DOI: }\fi
\providecommand{\bibinfo}[2]{#2}
\providecommand{\eprint}[2][]{\url{#2}}

\bibitem{Bray2018}
\bibinfo{author}{Bray, F.} \emph{et~al.}
\newblock \bibinfo{journal}{\bibinfo{title}{Global cancer statistics 2018:
  {GLOBOCAN} estimates of incidence and mortality worldwide for 36 cancers in
  185 countries}}.
\newblock {\emph{\JournalTitle{CA Cancer J Clin.}}}
  \textbf{\bibinfo{volume}{68}}, \bibinfo{pages}{394--424}
  (\bibinfo{year}{2018}).

\bibitem{kaminski2017increased}
\bibinfo{author}{Kaminski, M.~F.} \emph{et~al.}
\newblock \bibinfo{journal}{\bibinfo{title}{Increased rate of adenoma detection
  associates with reduced risk of colorectal cancer and death}}.
\newblock {\emph{\JournalTitle{Gastroenterology}}}
  \textbf{\bibinfo{volume}{153}}, \bibinfo{pages}{98--105}
  (\bibinfo{year}{2017}).

\bibitem{urban2018deep}
\bibinfo{author}{Urban, G.} \emph{et~al.}
\newblock \bibinfo{journal}{\bibinfo{title}{Deep learning localizes and
  identifies polyps in real time with 96\% accuracy in screening colonoscopy}}.
\newblock {\emph{\JournalTitle{Gastroenterology}}}
  \textbf{\bibinfo{volume}{155}}, \bibinfo{pages}{1069--1078}
  (\bibinfo{year}{2018}).

\bibitem{2019Polyp}
\bibinfo{author}{Qadir, H.~A.} \emph{et~al.}
\newblock \bibinfo{title}{Polyp detection and segmentation using mask r-cnn:
  Does a deeper feature extractor cnn always perform better?}
\newblock In \emph{\bibinfo{booktitle}{2019 13th International Symposium on
  Medical Information and Communication Technology (ISMICT)}},
  \bibinfo{pages}{1--6} (\bibinfo{year}{2019}).

\bibitem{zhang2019real}
\bibinfo{author}{Zhang, X.} \emph{et~al.}
\newblock \bibinfo{journal}{\bibinfo{title}{Real-time gastric polyp detection
  using convolutional neural networks}}.
\newblock {\emph{\JournalTitle{PloS one}}} \textbf{\bibinfo{volume}{14}},
  \bibinfo{pages}{e0214133} (\bibinfo{year}{2019}).

\bibitem{nguyen2019robust}
\bibinfo{author}{Nguyen, N.-Q.} \& \bibinfo{author}{Lee, S.-W.}
\newblock \bibinfo{journal}{\bibinfo{title}{Robust boundary segmentation in
  medical images using a consecutive deep encoder-decoder network}}.
\newblock {\emph{\JournalTitle{Ieee Access}}} \textbf{\bibinfo{volume}{7}},
  \bibinfo{pages}{33795--33808} (\bibinfo{year}{2019}).

\bibitem{guo2019giana}
\bibinfo{author}{Guo, Y.~B.} \& \bibinfo{author}{Matuszewski, B.}
\newblock \bibinfo{title}{Giana polyp segmentation with fully convolutional
  dilation neural networks}.
\newblock In \emph{\bibinfo{booktitle}{Proceedings of the 14th International
  Joint Conference on Computer Vision, Imaging and Computer Graphics Theory and
  Applications}}, \bibinfo{pages}{632--641}
  (\bibinfo{organization}{SCITEPRESS-Science and Technology Publications},
  \bibinfo{year}{2019}).

\bibitem{zhang2021transfuse}
\bibinfo{author}{Zhang, Y.}, \bibinfo{author}{Liu, H.} \& \bibinfo{author}{Hu,
  Q.}
\newblock \bibinfo{journal}{\bibinfo{title}{Transfuse: Fusing transformers and
  {CNNs} for medical image segmentation}}.
\newblock {\emph{\JournalTitle{arXiv preprint arXiv:2102.08005}}}
  (\bibinfo{year}{2021}).

\bibitem{ali2021_endoCV2020}
\bibinfo{author}{Ali, S.} \emph{et~al.}
\newblock \bibinfo{journal}{\bibinfo{title}{Deep learning for detection and
  segmentation of artefact and disease instances in gastrointestinal
  endoscopy}}.
\newblock {\emph{\JournalTitle{Medical Image Analysis}}}
  \textbf{\bibinfo{volume}{70}}, \bibinfo{pages}{102002},
  \doiprefix\url{10.1016/j.media.2021.102002} (\bibinfo{year}{2021}).

\bibitem{srivastava2021msrf}
\bibinfo{author}{Srivastava, A.} \emph{et~al.}
\newblock \bibinfo{journal}{\bibinfo{title}{{MSRF-Net: A Multi-Scale Residual
  Fusion Network for Biomedical Image Segmentation}}}.
\newblock {\emph{\JournalTitle{IEEE Journal of Biomedical and Health
  informatics}}}  (\bibinfo{year}{2021}).

\bibitem{Celik2021:MICCAI}
\bibinfo{author}{Celik, N.}, \bibinfo{author}{Ali, S.}, \bibinfo{author}{Gupta,
  S.}, \bibinfo{author}{Braden, B.} \& \bibinfo{author}{Rittscher, J.}
\newblock \bibinfo{title}{Endouda: A modality independent segmentation approach
  for endoscopy imaging}.
\newblock In \emph{\bibinfo{booktitle}{Medical Image Computing and Computer
  Assisted Intervention -- MICCAI 2021}}, \bibinfo{pages}{303--312}
  (\bibinfo{publisher}{Springer International Publishing},
  \bibinfo{year}{2021}).

\bibitem{silva2014toward}
\bibinfo{author}{Silva, J.}, \bibinfo{author}{Histace, A.},
  \bibinfo{author}{Romain, O.}, \bibinfo{author}{Dray, X.} \&
  \bibinfo{author}{Granado, B.}
\newblock \bibinfo{journal}{\bibinfo{title}{Toward embedded detection of polyps
  in wce images for early diagnosis of colorectal cancer}}.
\newblock {\emph{\JournalTitle{International Journal of Computer Assisted
  Radiology and Surgery}}} \textbf{\bibinfo{volume}{9}},
  \bibinfo{pages}{283--293} (\bibinfo{year}{2014}).

\bibitem{bernal2015wm}
\bibinfo{author}{Bernal, J.} \emph{et~al.}
\newblock \bibinfo{journal}{\bibinfo{title}{Wm-dova maps for accurate polyp
  highlighting in colonoscopy: Validation vs. saliency maps from physicians}}.
\newblock {\emph{\JournalTitle{Computerized Medical Imaging and Graphics}}}
  \textbf{\bibinfo{volume}{43}}, \bibinfo{pages}{99--111}
  (\bibinfo{year}{2015}).

\bibitem{borgli2020hyperkvasir}
\bibinfo{author}{Borgli, H.} \emph{et~al.}
\newblock \bibinfo{journal}{\bibinfo{title}{Hyperkvasir, a comprehensive
  multi-class image and video dataset for gastrointestinal endoscopy}}.
\newblock {\emph{\JournalTitle{Scientific Data}}} \textbf{\bibinfo{volume}{7}},
  \bibinfo{pages}{1--14} (\bibinfo{year}{2020}).

\bibitem{ali2020objective}
\bibinfo{author}{Ali, S.} \emph{et~al.}
\newblock \bibinfo{journal}{\bibinfo{title}{An objective comparison of
  detection and segmentation algorithms for artefacts in clinical endoscopy}}.
\newblock {\emph{\JournalTitle{Scientific reports}}}
  \textbf{\bibinfo{volume}{10}}, \bibinfo{pages}{1--15} (\bibinfo{year}{2020}).

\bibitem{ali2021polypgen}
\bibinfo{author}{Ali, S.} \emph{et~al.}
\newblock \bibinfo{journal}{\bibinfo{title}{{PolypGen:} a multi-center polyp
  detection and segmentation dataset for generalisability assessment}}.
\newblock {\emph{\JournalTitle{arXiv preprint arXiv:2106.04463}}}
  (\bibinfo{year}{2021}).

\bibitem{tajbakhsh2015automatic}
\bibinfo{author}{Tajbakhsh, N.}, \bibinfo{author}{Gurudu, S.~R.} \&
  \bibinfo{author}{Liang, J.}
\newblock \bibinfo{title}{Automatic polyp detection in colonoscopy videos using
  an ensemble of convolutional neural networks}.
\newblock In \emph{\bibinfo{booktitle}{2015 IEEE 12th International Symposium
  on Biomedical Imaging (ISBI)}}, \bibinfo{pages}{79--83}
  (\bibinfo{organization}{IEEE}, \bibinfo{year}{2015}).

\bibitem{park2016colonoscopic}
\bibinfo{author}{Park, S.~Y.} \& \bibinfo{author}{Sargent, D.}
\newblock \bibinfo{title}{Colonoscopic polyp detection using convolutional
  neural networks}.
\newblock In \emph{\bibinfo{booktitle}{Medical Imaging 2016: Computer-Aided
  Diagnosis}}, vol. \bibinfo{volume}{9785}, \bibinfo{pages}{978528}
  (\bibinfo{organization}{International Society for Optics and Photonics},
  \bibinfo{year}{2016}).

\bibitem{ribeiro2016colonic}
\bibinfo{author}{Ribeiro, E.}, \bibinfo{author}{Uhl, A.} \&
  \bibinfo{author}{H{\"a}fner, M.}
\newblock \bibinfo{title}{Colonic polyp classification with convolutional
  neural networks}.
\newblock In \emph{\bibinfo{booktitle}{2016 IEEE 29th International Symposium
  on Computer-Based Medical Systems (CBMS)}}, \bibinfo{pages}{253--258}
  (\bibinfo{organization}{IEEE}, \bibinfo{year}{2016}).

\bibitem{girshick2014rich}
\bibinfo{author}{Girshick, R.}, \bibinfo{author}{Donahue, J.},
  \bibinfo{author}{Darrell, T.} \& \bibinfo{author}{Malik, J.}
\newblock \bibinfo{title}{Rich feature hierarchies for accurate object
  detection and semantic segmentation}.
\newblock In \emph{\bibinfo{booktitle}{Proceedings of the IEEE conference on
  computer vision and pattern recognition}}, \bibinfo{pages}{580--587}
  (\bibinfo{year}{2014}).

\bibitem{girshick2015fast}
\bibinfo{author}{Girshick, R.}
\newblock \bibinfo{title}{Fast r-cnn}.
\newblock In \emph{\bibinfo{booktitle}{Proceedings of the IEEE international
  conference on computer vision}}, \bibinfo{pages}{1440--1448}
  (\bibinfo{year}{2015}).

\bibitem{ren2015faster}
\bibinfo{author}{Ren, S.}, \bibinfo{author}{He, K.}, \bibinfo{author}{Girshick,
  R.} \& \bibinfo{author}{Sun, J.}
\newblock \bibinfo{title}{Faster r-cnn: Towards real-time object detection with
  region proposal networks}.
\newblock In \emph{\bibinfo{booktitle}{Advances in neural information
  processing systems}}, \bibinfo{pages}{91--99} (\bibinfo{year}{2015}).

\bibitem{dai2016r}
\bibinfo{author}{Dai, J.}, \bibinfo{author}{Li, Y.}, \bibinfo{author}{He, K.}
  \& \bibinfo{author}{Sun, J.}
\newblock \bibinfo{title}{{R-FCN: Object detection via region-based fully
  convolutional networks}}.
\newblock In \emph{\bibinfo{booktitle}{Advances in neural information
  processing systems}}, \bibinfo{pages}{379--387} (\bibinfo{year}{2016}).

\bibitem{lin2017feature}
\bibinfo{author}{Lin, T.-Y.} \emph{et~al.}
\newblock \bibinfo{title}{Feature pyramid networks for object detection}.
\newblock In \emph{\bibinfo{booktitle}{Proceedings of the IEEE conference on
  computer vision and pattern recognition}}, \bibinfo{pages}{2117--2125}
  (\bibinfo{year}{2017}).

\bibitem{cai2018cascade}
\bibinfo{author}{Cai, Z.} \& \bibinfo{author}{Vasconcelos, N.}
\newblock \bibinfo{title}{Cascade r-cnn: Delving into high quality object
  detection}.
\newblock In \emph{\bibinfo{booktitle}{Proceedings of the IEEE Conference on
  Computer Vision and Pattern Recognition}}, \bibinfo{pages}{6154--6162}
  (\bibinfo{year}{2018}).

\bibitem{liu2016ssd}
\bibinfo{author}{Liu, W.} \emph{et~al.}
\newblock \bibinfo{title}{Ssd: Single shot multibox detector}.
\newblock In \emph{\bibinfo{booktitle}{European conference on computer
  vision}}, \bibinfo{pages}{21--37} (\bibinfo{organization}{Springer},
  \bibinfo{year}{2016}).

\bibitem{redmon2016you}
\bibinfo{author}{Redmon, J.}, \bibinfo{author}{Divvala, S.},
  \bibinfo{author}{Girshick, R.} \& \bibinfo{author}{Farhadi, A.}
\newblock \bibinfo{title}{You only look once: Unified, real-time object
  detection}.
\newblock In \emph{\bibinfo{booktitle}{Proceedings of the IEEE conference on
  computer vision and pattern recognition}}, \bibinfo{pages}{779--788}
  (\bibinfo{year}{2016}).

\bibitem{lin2017focal}
\bibinfo{author}{Lin, T.-Y.}, \bibinfo{author}{Goyal, P.},
  \bibinfo{author}{Girshick, R.}, \bibinfo{author}{He, K.} \&
  \bibinfo{author}{Doll{\'a}r, P.}
\newblock \bibinfo{title}{Focal loss for dense object detection}.
\newblock In \emph{\bibinfo{booktitle}{Proceedings of the IEEE international
  conference on computer vision}}, \bibinfo{pages}{2980--2988}
  (\bibinfo{year}{2017}).

\bibitem{tan2020efficientdet}
\bibinfo{author}{Tan, M.}, \bibinfo{author}{Pang, R.} \& \bibinfo{author}{Le,
  Q.~V.}
\newblock \bibinfo{title}{Efficientdet: Scalable and efficient object
  detection}.
\newblock In \emph{\bibinfo{booktitle}{Proceedings of the IEEE/CVF conference
  on computer vision and pattern recognition}}, \bibinfo{pages}{10781--10790}
  (\bibinfo{year}{2020}).

\bibitem{younghak2017}
\bibinfo{author}{Y., S.}, \bibinfo{author}{H.~A., Q.}, \bibinfo{author}{L.,
  A.}, \bibinfo{author}{J., B.} \& \bibinfo{author}{I., B.}
\newblock \bibinfo{journal}{\bibinfo{title}{Automatic colon polyp detection
  using region based deep cnn and post learning approaches}}.
\newblock {\emph{\JournalTitle{IEEE Access}}} \textbf{\bibinfo{volume}{6}},
  \bibinfo{pages}{40950--40962} (\bibinfo{year}{2018}).

\bibitem{he2017mask}
\bibinfo{author}{He, K.}, \bibinfo{author}{Gkioxari, G.},
  \bibinfo{author}{Doll{\'a}r, P.} \& \bibinfo{author}{Girshick, R.}
\newblock \bibinfo{title}{Mask r-cnn}.
\newblock In \emph{\bibinfo{booktitle}{Proceedings of the IEEE international
  conference on computer vision}}, \bibinfo{pages}{2961--2969}
  (\bibinfo{year}{2017}).

\bibitem{he2016deep}
\bibinfo{author}{He, K.}, \bibinfo{author}{Zhang, X.}, \bibinfo{author}{Ren,
  S.} \& \bibinfo{author}{Sun, J.}
\newblock \bibinfo{title}{Deep residual learning for image recognition}.
\newblock In \emph{\bibinfo{booktitle}{Proceedings of the IEEE conference on
  computer vision and pattern recognition}}, \bibinfo{pages}{770--778}
  (\bibinfo{year}{2016}).

\bibitem{2016Inception}
\bibinfo{author}{Szegedy, C.}, \bibinfo{author}{Ioffe, S.},
  \bibinfo{author}{Vanhoucke, V.} \& \bibinfo{author}{Alemi, A.~A.}
\newblock \bibinfo{title}{Inception-v4, inception-resnet and the impact of
  residual connections on learning}.
\newblock In \emph{\bibinfo{booktitle}{Thirty-first AAAI conference on
  artificial intelligence}} (\bibinfo{year}{2017}).

\bibitem{lee2020real}
\bibinfo{author}{Lee, J.~Y.} \emph{et~al.}
\newblock \bibinfo{journal}{\bibinfo{title}{Real-time detection of colon polyps
  during colonoscopy using deep learning: systematic validation with four
  independent datasets}}.
\newblock {\emph{\JournalTitle{Scientific reports}}}
  \textbf{\bibinfo{volume}{10}}, \bibinfo{pages}{1--9} (\bibinfo{year}{2020}).

\bibitem{redmon2017yolo9000}
\bibinfo{author}{Redmon, J.} \& \bibinfo{author}{Farhadi, A.}
\newblock \bibinfo{title}{Yolo9000: better, faster, stronger}.
\newblock In \emph{\bibinfo{booktitle}{Proceedings of the IEEE conference on
  computer vision and pattern recognition}}, \bibinfo{pages}{7263--7271}
  (\bibinfo{year}{2017}).

\bibitem{zhang2018polyp}
\bibinfo{author}{Zhang, R.}, \bibinfo{author}{Zheng, Y.},
  \bibinfo{author}{Poon, C.~C.}, \bibinfo{author}{Shen, D.} \&
  \bibinfo{author}{Lau, J.~Y.}
\newblock \bibinfo{journal}{\bibinfo{title}{Polyp detection during colonoscopy
  using a regression-based convolutional neural network with a tracker}}.
\newblock {\emph{\JournalTitle{Pattern recognition}}}
  \textbf{\bibinfo{volume}{83}}, \bibinfo{pages}{209--219}
  (\bibinfo{year}{2018}).

\bibitem{redmon2018yolov3}
\bibinfo{author}{Farhadi, A.} \& \bibinfo{author}{Redmon, J.}
\newblock \bibinfo{title}{Yolov3: An incremental improvement}.
\newblock In \emph{\bibinfo{booktitle}{Computer Vision and Pattern
  Recognition}}, \bibinfo{pages}{1804--2767} (\bibinfo{organization}{Springer
  Berlin/Heidelberg, Germany}, \bibinfo{year}{2018}).

\bibitem{law2018cornernet}
\bibinfo{author}{Law, H.} \& \bibinfo{author}{Deng, J.}
\newblock \bibinfo{title}{Cornernet: Detecting objects as paired keypoints}.
\newblock In \emph{\bibinfo{booktitle}{Proceedings of the European Conference
  on Computer Vision (ECCV)}}, \bibinfo{pages}{734--750}
  (\bibinfo{year}{2018}).

\bibitem{zhou2019bottom}
\bibinfo{author}{Zhou, X.}, \bibinfo{author}{Zhuo, J.} \&
  \bibinfo{author}{Krahenbuhl, P.}
\newblock \bibinfo{title}{Bottom-up object detection by grouping extreme and
  center points}.
\newblock In \emph{\bibinfo{booktitle}{Proceedings of the IEEE Conference on
  Computer Vision and Pattern Recognition}}, \bibinfo{pages}{850--859}
  (\bibinfo{year}{2019}).

\bibitem{2019Objects}
\bibinfo{author}{Zhou, X.}, \bibinfo{author}{Wang, D.} \&
  \bibinfo{author}{Kr{\"a}henb{\"u}hl, P.}
\newblock \bibinfo{journal}{\bibinfo{title}{Objects as points}}.
\newblock {\emph{\JournalTitle{arXiv preprint arXiv:1904.07850}}}
  (\bibinfo{year}{2019}).

\bibitem{wang2019afp}
\bibinfo{author}{Wang, D.} \emph{et~al.}
\newblock \bibinfo{title}{{AFP-Net: Realtime Anchor-Free Polyp Detection in
  Colonoscopy}}.
\newblock In \emph{\bibinfo{booktitle}{2019 IEEE 31st International Conference
  on Tools with Artificial Intelligence (ICTAI)}}, \bibinfo{pages}{636--643}
  (\bibinfo{organization}{IEEE}, \bibinfo{year}{2019}).

\bibitem{7840040}
\bibinfo{author}{Bernal, J.} \emph{et~al.}
\newblock \bibinfo{journal}{\bibinfo{title}{Comparative validation of polyp
  detection methods in video colonoscopy: Results from the miccai 2015
  endoscopic vision challenge}}.
\newblock {\emph{\JournalTitle{IEEE Transactions on Medical Imaging}}}
  \textbf{\bibinfo{volume}{36}}, \bibinfo{pages}{1231--1249},
  \doiprefix\url{10.1109/TMI.2017.2664042} (\bibinfo{year}{2017}).

\bibitem{ross2020robust}
\bibinfo{author}{Ross, T.} \emph{et~al.}
\newblock \bibinfo{journal}{\bibinfo{title}{Robust medical instrument
  segmentation challenge 2019}}.
\newblock {\emph{\JournalTitle{arXiv preprint arXiv:2003.10299}}}
  (\bibinfo{year}{2020}).

\bibitem{brandao2017fully}
\bibinfo{author}{Brandao, P.} \emph{et~al.}
\newblock \bibinfo{title}{Fully convolutional neural networks for polyp
  segmentation in colonoscopy}.
\newblock In \emph{\bibinfo{booktitle}{Medical Imaging 2017: Computer-Aided
  Diagnosis}}, vol. \bibinfo{volume}{10134}, \bibinfo{pages}{101340F}
  (\bibinfo{organization}{International Society for Optics and Photonics},
  \bibinfo{year}{2017}).

\bibitem{zhang2017automated}
\bibinfo{author}{Zhang, L.}, \bibinfo{author}{Dolwani, S.} \&
  \bibinfo{author}{Ye, X.}
\newblock \bibinfo{title}{Automated polyp segmentation in colonoscopy frames
  using fully convolutional neural network and textons}.
\newblock In \emph{\bibinfo{booktitle}{Annual Conference on Medical Image
  Understanding and Analysis}}, \bibinfo{pages}{707--717}
  (\bibinfo{organization}{Springer}, \bibinfo{year}{2017}).

\bibitem{akbari2018polyp}
\bibinfo{author}{Akbari, M.} \emph{et~al.}
\newblock \bibinfo{title}{Polyp segmentation in colonoscopy images using fully
  convolutional network}.
\newblock In \emph{\bibinfo{booktitle}{2018 40th Annual International
  Conference of the IEEE Engineering in Medicine and Biology Society (EMBC)}},
  \bibinfo{pages}{69--72} (\bibinfo{organization}{IEEE}, \bibinfo{year}{2018}).

\bibitem{zhou2019unet++}
\bibinfo{author}{Zhou, Z.}, \bibinfo{author}{Siddiquee, M. M.~R.},
  \bibinfo{author}{Tajbakhsh, N.} \& \bibinfo{author}{Liang, J.}
\newblock \bibinfo{journal}{\bibinfo{title}{{UNet++:} redesigning skip
  connections to exploit multiscale features in image segmentation}}.
\newblock {\emph{\JournalTitle{IEEE Transactions on Medical Imaging}}}
  \textbf{\bibinfo{volume}{39}}, \bibinfo{pages}{1856--1867}
  (\bibinfo{year}{2019}).

\bibitem{fan2020pranet}
\bibinfo{author}{Fan, D.-P.} \emph{et~al.}
\newblock \bibinfo{title}{Pranet: Parallel reverse attention network for polyp
  segmentation}.
\newblock In \emph{\bibinfo{booktitle}{International Conference on Medical
  Image Computing and Computer-Assisted Intervention}},
  \bibinfo{pages}{263--273} (\bibinfo{organization}{Springer},
  \bibinfo{year}{2020}).

\bibitem{mahmud2021polypsegnet}
\bibinfo{author}{Mahmud, T.}, \bibinfo{author}{Paul, B.} \&
  \bibinfo{author}{Fattah, S.~A.}
\newblock \bibinfo{journal}{\bibinfo{title}{{PolypSegNet:} a modified
  encoder-decoder architecture for automated polyp segmentation from
  colonoscopy images}}.
\newblock {\emph{\JournalTitle{Computers in Biology and Medicine}}}
  \textbf{\bibinfo{volume}{128}}, \bibinfo{pages}{104119}
  (\bibinfo{year}{2021}).

\bibitem{huang2021hardnet}
\bibinfo{author}{Huang, C.-H.}, \bibinfo{author}{Wu, H.-Y.} \&
  \bibinfo{author}{Lin, Y.-L.}
\newblock \bibinfo{journal}{\bibinfo{title}{{HarDNet-MSEG: A Simple
  Encoder-Decoder Polyp Segmentation Neural Network that Achieves over 0.9 Mean
  Dice and 86 FPS}}}.
\newblock {\emph{\JournalTitle{arXiv preprint arXiv:2101.07172}}}
  (\bibinfo{year}{2021}).

\bibitem{wang2020deep}
\bibinfo{author}{Wang, J.} \emph{et~al.}
\newblock \bibinfo{journal}{\bibinfo{title}{Deep high-resolution representation
  learning for visual recognition}}.
\newblock {\emph{\JournalTitle{IEEE transactions on pattern analysis and
  machine intelligence}}}  (\bibinfo{year}{2020}).

\bibitem{ji2021progressively}
\bibinfo{author}{Ji, G.-P.} \emph{et~al.}
\newblock \bibinfo{journal}{\bibinfo{title}{Progressively normalized
  self-attention network for video polyp segmentation}}.
\newblock {\emph{\JournalTitle{arXiv preprint arXiv:2105.08468}}}
  (\bibinfo{year}{2021}).

\bibitem{Jia2020}
\bibinfo{author}{Jia, X.} \emph{et~al.}
\newblock \bibinfo{journal}{\bibinfo{title}{{Automatic Polyp Recognition in
  Colonoscopy Images Using Deep Learning and Two-Stage Pyramidal Feature
  Prediction}}}.
\newblock {\emph{\JournalTitle{IEEE Transactions on Automation Science and
  Engineering}}} \textbf{\bibinfo{volume}{17}} (\bibinfo{year}{2020}).

\bibitem{Guo2019}
\bibinfo{author}{Guo, X.} \emph{et~al.}
\newblock \bibinfo{journal}{\bibinfo{title}{{Automated polyp segmentation for
  colonoscopy images: A method based on convolutional neural networks and
  ensemble learning}}}.
\newblock {\emph{\JournalTitle{Medical Physics}}}
  \textbf{\bibinfo{volume}{46}}, \bibinfo{pages}{5666--5676}
  (\bibinfo{year}{2019}).

\bibitem{Zhao2017}
\bibinfo{author}{Zhao, H.}, \bibinfo{author}{Shi, J.}, \bibinfo{author}{Qi,
  X.}, \bibinfo{author}{Wang, X.} \& \bibinfo{author}{Jia, J.}
\newblock \bibinfo{journal}{\bibinfo{title}{{Pyramid scene parsing network}}}.
\newblock {\emph{\JournalTitle{Proceedings - 30th IEEE Conference on Computer
  Vision and Pattern Recognition, CVPR 2017}}}
  \doiprefix\url{10.1109/CVPR.2017.660} (\bibinfo{year}{2017}).

\bibitem{badrinarayanan2017segnet}
\bibinfo{author}{Badrinarayanan, V.}, \bibinfo{author}{Kendall, A.} \&
  \bibinfo{author}{Cipolla, R.}
\newblock \bibinfo{journal}{\bibinfo{title}{Segnet: A deep convolutional
  encoder-decoder architecture for image segmentation}}.
\newblock {\emph{\JournalTitle{IEEE transactions on pattern analysis and
  machine intelligence}}} \textbf{\bibinfo{volume}{39}},
  \bibinfo{pages}{2481--2495} (\bibinfo{year}{2017}).

\bibitem{ronneberger2015u}
\bibinfo{author}{Ronneberger, O.}, \bibinfo{author}{Fischer, P.} \&
  \bibinfo{author}{Brox, T.}
\newblock \bibinfo{title}{U-net: Convolutional networks for biomedical image
  segmentation}.
\newblock In \emph{\bibinfo{booktitle}{International Conference on Medical
  image computing and computer-assisted intervention}},
  \bibinfo{pages}{234--241} (\bibinfo{organization}{Springer},
  \bibinfo{year}{2015}).

\bibitem{sun2019colorectal}
\bibinfo{author}{Sun, X.}, \bibinfo{author}{Zhang, P.}, \bibinfo{author}{Wang,
  D.}, \bibinfo{author}{Cao, Y.} \& \bibinfo{author}{Liu, B.}
\newblock \bibinfo{title}{Colorectal polyp segmentation by u-net with dilation
  convolution}.
\newblock In \emph{\bibinfo{booktitle}{2019 18th IEEE International Conference
  On Machine Learning And Applications (ICMLA)}}, \bibinfo{pages}{851--858}
  (\bibinfo{organization}{IEEE}, \bibinfo{year}{2019}).

\bibitem{safarov2021denseunet}
\bibinfo{author}{Safarov, S.} \& \bibinfo{author}{Whangbo, T.~K.}
\newblock \bibinfo{journal}{\bibinfo{title}{{A-DenseUNet:} adaptive densely
  connected unet for polyp segmentation in colonoscopy images with atrous
  convolution}}.
\newblock {\emph{\JournalTitle{Sensors}}} \textbf{\bibinfo{volume}{21}},
  \bibinfo{pages}{1441} (\bibinfo{year}{2021}).

\bibitem{wang2019nested}
\bibinfo{author}{Wang, L.} \emph{et~al.}
\newblock \bibinfo{journal}{\bibinfo{title}{{Nested Dilation Network (NDN) for
  Multi-Task Medical Image Segmentation}}}.
\newblock {\emph{\JournalTitle{IEEE Access}}} \textbf{\bibinfo{volume}{7}},
  \bibinfo{pages}{44676--44685}, \doiprefix\url{10.1109/ACCESS.2019.2908386}
  (\bibinfo{year}{2019}).

\bibitem{lin2014microsoft}
\bibinfo{author}{Lin, T.-Y.} \emph{et~al.}
\newblock \bibinfo{title}{{Microsoft COCO: Common objects in context}}.
\newblock In \emph{\bibinfo{booktitle}{European conference on computer
  vision}}, \bibinfo{pages}{740--755} (\bibinfo{year}{2014}).

\bibitem{Ali2021:EndoCVProc}
\bibinfo{editor}{Ali, S.}, \bibinfo{editor}{Ghatwary, N.~M.},
  \bibinfo{editor}{Jha, D.} \& \bibinfo{editor}{Halvorsen, P.} (eds.).
\newblock \emph{\bibinfo{title}{Proceedings of the 3rd International Workshop
  and Challenge on Computer Vision in Endoscopy (EndoCV 2021) co-located with
  with the 18th {IEEE} International Symposium on Biomedical Imaging {(ISBI}
  2021), Nice, France, April 13, 2021}}, vol. \bibinfo{volume}{2886} of
  \emph{\bibinfo{series}{{CEUR} Workshop Proceedings}}
  (\bibinfo{publisher}{CEUR-WS.org}, \bibinfo{year}{2021}).

\bibitem{lia2021joint}
\bibinfo{author}{Wuyang, L.} \emph{et~al.}
\newblock \bibinfo{title}{Joint polyp detection and segmentation with
  heterogeneous endoscopic data}.
\newblock In \emph{\bibinfo{booktitle}{Proceedings of the 3rd International
  Workshop and Challenge on Computer Vision in Endoscopy (EndoCV 2021)
  co-located with with the 18th {IEEE}International Symposium on Biomedical
  Imaging {(ISBI} 2021), Nice, France, April 13, 2021}}, vol.
  \bibinfo{volume}{2886}, \bibinfo{pages}{69--79}
  (\bibinfo{publisher}{CEUR-WS.org}, \bibinfo{year}{2021}).

\bibitem{honga2021deep}
\bibinfo{author}{Honga, A.}, \bibinfo{author}{Leeb, G.}, \bibinfo{author}{Leec,
  H.}, \bibinfo{author}{Seod, J.} \& \bibinfo{author}{Yeoe, D.}
\newblock \bibinfo{title}{Deep learning model generalization with ensemble in
  endoscopic images}.
\newblock In \emph{\bibinfo{booktitle}{Proceedings of the 3rd International
  Workshop and Challenge on Computer Vision in Endoscopy (EndoCV 2021)
  co-located with with the 18th {IEEE}International Symposium on Biomedical
  Imaging {(ISBI} 2021), Nice, France, April 13, 2021}}, vol.
  \bibinfo{volume}{2886}, \bibinfo{pages}{80--89} (\bibinfo{year}{2021}).

\bibitem{gana2021detection}
\bibinfo{author}{Gana, T.}, \bibinfo{author}{Zhaa, Z.}, \bibinfo{author}{Hua,
  C.} \& \bibinfo{author}{Jina, Z.}
\newblock \bibinfo{title}{Detection of polyps during colonoscopy procedure
  using yolov5 network}.
\newblock In \emph{\bibinfo{booktitle}{Proceedings of the 3rd International
  Workshop and Challenge on Computer Vision in Endoscopy (EndoCV 2021)
  co-located with with the 18th {IEEE}International Symposium on Biomedical
  Imaging {(ISBI} 2021), Nice, France, April 13, 2021}}, vol.
  \bibinfo{volume}{2886}, \bibinfo{pages}{101--110}
  (\bibinfo{publisher}{CEUR-WS.org}, \bibinfo{year}{2021}).

\bibitem{polat2021polyp}
\bibinfo{author}{Polat, G.}, \bibinfo{author}{Isik-Polat, E.},
  \bibinfo{author}{Kayabay, K.} \& \bibinfo{author}{Temizel, A.}
\newblock \bibinfo{title}{Polyp detection in colonoscopy images using deep
  learning and bootstrap aggregation}.
\newblock In \emph{\bibinfo{booktitle}{Proceedings of the 3rd International
  Workshop and Challenge on Computer Vision in Endoscopy (EndoCV 2021)
  co-located with with the 18th {IEEE}International Symposium on Biomedical
  Imaging {(ISBI} 2021), Nice, France, April 13, 2021}}, vol.
  \bibinfo{volume}{2886}, \bibinfo{pages}{90--100}
  (\bibinfo{publisher}{CEUR-WS.org}, \bibinfo{year}{2021}).

\bibitem{tian2019fcos}
\bibinfo{author}{Tian, Z.}, \bibinfo{author}{Shen, C.}, \bibinfo{author}{Chen,
  H.} \& \bibinfo{author}{He, T.}
\newblock \bibinfo{title}{Fcos: Fully convolutional one-stage object
  detection}.
\newblock In \emph{\bibinfo{booktitle}{Proceedings of the IEEE/CVF
  international conference on computer vision}}, \bibinfo{pages}{9627--9636}
  (\bibinfo{year}{2019}).

\bibitem{solovyev2021weighted}
\bibinfo{author}{Solovyev, R.}, \bibinfo{author}{Wang, W.} \&
  \bibinfo{author}{Gabruseva, T.}
\newblock \bibinfo{journal}{\bibinfo{title}{Weighted boxes fusion: Ensembling
  boxes from different object detection models}}.
\newblock {\emph{\JournalTitle{Image and Vision Computing}}}
  \textbf{\bibinfo{volume}{107}}, \bibinfo{pages}{104117},
  \doiprefix\url{https://doi.org/10.1016/j.imavis.2021.104117}
  (\bibinfo{year}{2021}).

\bibitem{yolov5}
\bibinfo{author}{Jocher, G.}
\newblock \bibinfo{title}{{YoloV5}}.
\newblock \bibinfo{howpublished}{https://github.com/ultralytics/yolov5}
  (\bibinfo{year}{2021}).

\bibitem{galdrana2021multi}
\bibinfo{author}{Galdran, A.}, \bibinfo{author}{Carneiro, G.} \&
  \bibinfo{author}{Ballester, M. {\'{A}}.~G.}
\newblock \bibinfo{title}{Multi-center polyp segmentation withdouble
  encoder-decoder networks}.
\newblock In \bibinfo{editor}{Ali, S.}, \bibinfo{editor}{Ghatwary, N.~M.},
  \bibinfo{editor}{Jha, D.} \& \bibinfo{editor}{Halvorsen, P.} (eds.)
  \emph{\bibinfo{booktitle}{Proceedings of the 3rd International Workshop and
  Challenge on Computer Vision in Endoscopy (EndoCV 2021) co-located with with
  the 18th {IEEE}International Symposium on Biomedical Imaging {(ISBI} 2021),
  Nice, France, April 13, 2021}}, vol. \bibinfo{volume}{2886},
  \bibinfo{pages}{9--16} (\bibinfo{publisher}{CEUR-WS.org},
  \bibinfo{year}{2021}).

\bibitem{thambawita2021divergentnets}
\bibinfo{author}{Thambawita, V.}, \bibinfo{author}{Hicks, S.~A.},
  \bibinfo{author}{Halvorsen, P.} \& \bibinfo{author}{Riegler, M.~A.}
\newblock \bibinfo{title}{Divergentnets: Medical image segmentation by network
  ensemble}.
\newblock In \emph{\bibinfo{booktitle}{Proceedings of the 3rd International
  Workshop and Challenge on Computer Vision in Endoscopy (EndoCV 2021)
  co-located with with the 18th {IEEE} International Symposium on Biomedical
  Imaging {(ISBI} 2021), Nice, France, April 13, 2021}}, vol.
  \bibinfo{volume}{2886}, \bibinfo{pages}{27--38} (\bibinfo{year}{2021}).

\bibitem{ghimirea2021augmentation}
\bibinfo{author}{Ghimirea, R.}, \bibinfo{author}{Poudelb, S.} \&
  \bibinfo{author}{Leec, S.-W.}
\newblock \bibinfo{title}{An augmentation strategy with lightweight network for
  polyp segmentation}.
\newblock In \emph{\bibinfo{booktitle}{Proceedings of the 3rd International
  Workshop and Challenge on Computer Vision in Endoscopy (EndoCV 2021)
  co-located with with the 18th {IEEE}International Symposium on Biomedical
  Imaging {(ISBI} 2021), Nice, France, April 13, 2021}}, vol.
  \bibinfo{volume}{2886}, \bibinfo{pages}{39--48} (\bibinfo{year}{2021}).

\bibitem{yua2021parallel}
\bibinfo{author}{Yua, C.}, \bibinfo{author}{Yana, J.} \& \bibinfo{author}{Lia,
  X.}
\newblock \bibinfo{title}{Parallel res2net-based network with reverse attention
  for polyp segmentation}.
\newblock In \emph{\bibinfo{booktitle}{Proceedings of the 3rd International
  Workshop and Challenge on Computer Vision in Endoscopy (EndoCV 2021)
  co-located with with the 18th {IEEE}International Symposium on Biomedical
  Imaging {(ISBI} 2021), Nice, France, April 13, 2021}}, vol.
  \bibinfo{volume}{2886}, \bibinfo{pages}{17--26}
  (\bibinfo{publisher}{CEUR-WS.org}, \bibinfo{year}{2021}).

\bibitem{haithamia2021embedded}
\bibinfo{author}{Haithamia, M.}, \bibinfo{author}{Ahmeda, A.},
  \bibinfo{author}{Liaoa, I.~Y.} \& \bibinfo{author}{Jalaba, H.}
\newblock \bibinfo{title}{An embedded recurrent neural network-based model for
  endoscopic semantic segmentation}.
\newblock In \emph{\bibinfo{booktitle}{Proceedings of the 3rd International
  Workshop and Challenge on Computer Vision in Endoscopy (EndoCV 2021)
  co-located with with the 18th {IEEE}International Symposium on Biomedical
  Imaging {(ISBI} 2021), Nice, France, April 13, 2021}}, vol.
  \bibinfo{volume}{2886}, \bibinfo{pages}{49--58}
  (\bibinfo{publisher}{CEUR-WS.org}, \bibinfo{year}{2021}).

\bibitem{tomara2021improving}
\bibinfo{author}{Tomar, N.~K.}, \bibinfo{author}{Ibtehaz, N.},
  \bibinfo{author}{Jha, D.}, \bibinfo{author}{Halvorsen, P.} \&
  \bibinfo{author}{Ali, S.}
\newblock \bibinfo{title}{Improving generalizability in polyp segmentation
  using ensemble convolutional neural network}.
\newblock In \emph{\bibinfo{booktitle}{Proceedings of the 3rd International
  Workshop and Challenge on Computer Vision in Endoscopy (EndoCV 2021)
  co-located with with the 18th {IEEE}International Symposium on Biomedical
  Imaging {(ISBI} 2021), Nice, France, April 13, 2021}}, vol.
  \bibinfo{volume}{2886}, \bibinfo{pages}{49--58} (\bibinfo{year}{2021}).

\bibitem{galdran2021double}
\bibinfo{author}{Galdran, A.}, \bibinfo{author}{Carneiro, G.} \&
  \bibinfo{author}{Ballester, M. A.~G.}
\newblock \bibinfo{title}{Double encoder-decoder networks for gastrointestinal
  polyp segmentation}.
\newblock In \emph{\bibinfo{booktitle}{Pattern Recognition. ICPR International
  Workshops and Challenges: Virtual Event, January 10--15, 2021, Proceedings,
  Part I}}, \bibinfo{pages}{293--307} (\bibinfo{organization}{Springer
  International Publishing}, \bibinfo{year}{2021}).

\bibitem{chen2017rethinking}
\bibinfo{author}{Chen, L.-C.}, \bibinfo{author}{Papandreou, G.},
  \bibinfo{author}{Schroff, F.} \& \bibinfo{author}{Adam, H.}
\newblock \bibinfo{journal}{\bibinfo{title}{Rethinking atrous convolution for
  semantic image segmentation}}.
\newblock {\emph{\JournalTitle{arXiv preprint arXiv:1706.05587}}}
  (\bibinfo{year}{2017}).

\bibitem{chen2018encoder}
\bibinfo{author}{Chen, L.-C.}, \bibinfo{author}{Zhu, Y.},
  \bibinfo{author}{Papandreou, G.}, \bibinfo{author}{Schroff, F.} \&
  \bibinfo{author}{Adam, H.}
\newblock \bibinfo{title}{Encoder-decoder with atrous separable convolution for
  semantic image segmentation}.
\newblock In \emph{\bibinfo{booktitle}{Proceedings of the European conference
  on computer vision (ECCV)}}, \bibinfo{pages}{801--818}
  (\bibinfo{year}{2018}).

\bibitem{cho2014learning}
\bibinfo{author}{Cho, K.} \emph{et~al.}
\newblock \bibinfo{journal}{\bibinfo{title}{Learning phrase representations
  using rnn encoder-decoder for statistical machine translation}}.
\newblock {\emph{\JournalTitle{arXiv preprint arXiv:1406.1078}}}
  (\bibinfo{year}{2014}).

\bibitem{ibtehaz2020multiresunet}
\bibinfo{author}{Ibtehaz, N.} \& \bibinfo{author}{Rahman, M.~S.}
\newblock \bibinfo{journal}{\bibinfo{title}{{MultiResUNet: Rethinking the U-Net
  architecture for multimodal biomedical image segmentation}}}.
\newblock {\emph{\JournalTitle{Neural Networks}}}
  \textbf{\bibinfo{volume}{121}}, \bibinfo{pages}{74--87}
  (\bibinfo{year}{2020}).

\bibitem{YOLOv4-Alex2020}
\bibinfo{author}{Bochkovskiy, A.}, \bibinfo{author}{Wang, C.} \&
  \bibinfo{author}{Liao, H.~M.}
\newblock \bibinfo{journal}{\bibinfo{title}{Yolov4: Optimal speed and accuracy
  of object detection}}.
\newblock {\emph{\JournalTitle{CoRR}}}
  \textbf{\bibinfo{volume}{abs/2004.10934}} (\bibinfo{year}{2020}).
\newblock \eprint{2004.10934}.

\bibitem{chen2017deeplab}
\bibinfo{author}{Chen, L.-C.}, \bibinfo{author}{Papandreou, G.},
  \bibinfo{author}{Kokkinos, I.}, \bibinfo{author}{Murphy, K.} \&
  \bibinfo{author}{Yuille, A.~L.}
\newblock \bibinfo{journal}{\bibinfo{title}{Deeplab: Semantic image
  segmentation with deep convolutional nets, atrous convolution, and fully
  connected crfs}}.
\newblock {\emph{\JournalTitle{IEEE transactions on pattern analysis and
  machine intelligence}}} \textbf{\bibinfo{volume}{40}},
  \bibinfo{pages}{834--848} (\bibinfo{year}{2017}).

\bibitem{zhao2017pyramid}
\bibinfo{author}{Zhao, H.}, \bibinfo{author}{Shi, J.}, \bibinfo{author}{Qi,
  X.}, \bibinfo{author}{Wang, X.} \& \bibinfo{author}{Jia, J.}
\newblock \bibinfo{title}{Pyramid scene parsing network}.
\newblock In \emph{\bibinfo{booktitle}{Proceedings of the IEEE conference on
  computer vision and pattern recognition}}, \bibinfo{pages}{2881--2890}
  (\bibinfo{year}{2017}).

\bibitem{long2015fully}
\bibinfo{author}{Long, J.}, \bibinfo{author}{Shelhamer, E.} \&
  \bibinfo{author}{Darrell, T.}
\newblock \bibinfo{title}{Fully convolutional networks for semantic
  segmentation}.
\newblock In \emph{\bibinfo{booktitle}{Proceedings of the IEEE conference on
  computer vision and pattern recognition}}, \bibinfo{pages}{3431--3440}
  (\bibinfo{year}{2015}).

\bibitem{zhang2018road}
\bibinfo{author}{Zhang, Z.}, \bibinfo{author}{Liu, Q.} \&
  \bibinfo{author}{Wang, Y.}
\newblock \bibinfo{journal}{\bibinfo{title}{Road extraction by deep residual
  u-net}}.
\newblock {\emph{\JournalTitle{IEEE Geoscience and Remote Sensing Letters}}}
  \textbf{\bibinfo{volume}{15}}, \bibinfo{pages}{749--753}
  (\bibinfo{year}{2018}).

\bibitem{Wiesenfarth2021}
\bibinfo{author}{Wiesenfarth, M.} \emph{et~al.}
\newblock \bibinfo{journal}{\bibinfo{title}{Methods and open-source toolkit for
  analyzing and visualizing challenge results}}.
\newblock {\emph{\JournalTitle{Scientific Reports}}}
  \textbf{\bibinfo{volume}{11}}, \bibinfo{pages}{2369},
  \doiprefix\url{10.1038/s41598-021-82017-6} (\bibinfo{year}{2021}).

\bibitem{YongyiCVPR2017:DISCUSSION}
\bibinfo{author}{Lu, Y.}, \bibinfo{author}{Lu, C.} \& \bibinfo{author}{Tang,
  C.-K.}
\newblock \bibinfo{title}{Online video object detection using association
  lstm}.
\newblock In \emph{\bibinfo{booktitle}{2017 IEEE International Conference on
  Computer Vision (ICCV)}}, \bibinfo{pages}{2363--2371},
  \doiprefix\url{10.1109/ICCV.2017.257} (\bibinfo{year}{2017}).

\bibitem{bernal2012towards}
\bibinfo{author}{Bernal, J.}, \bibinfo{author}{S{\'a}nchez, J.} \&
  \bibinfo{author}{Vilarino, F.}
\newblock \bibinfo{journal}{\bibinfo{title}{Towards automatic polyp detection
  with a polyp appearance model}}.
\newblock {\emph{\JournalTitle{Pattern Recognition}}}
  \textbf{\bibinfo{volume}{45}}, \bibinfo{pages}{3166--3182}
  (\bibinfo{year}{2012}).

\bibitem{bernal2017miccai}
\bibinfo{author}{Bernal, J.} \& \bibinfo{author}{Aymeric, H.}
\newblock \bibinfo{journal}{\bibinfo{title}{Miccai endoscopic vision challenge
  polyp detection and segmentation}}.
\newblock {\emph{\JournalTitle{Web-page of the 2017 Endoscopic Vision
  Challenge}}}  (\bibinfo{year}{2017}).

\bibitem{ali2020endoscopy}
\bibinfo{author}{Ali, S.} \emph{et~al.}
\newblock \bibinfo{journal}{\bibinfo{title}{Endoscopy disease detection
  challenge 2020}}.
\newblock {\emph{\JournalTitle{arXiv preprint arXiv:2003.03376}}}
  (\bibinfo{year}{2020}).

\bibitem{tajbakhsh2015automated}
\bibinfo{author}{Tajbakhsh, N.}, \bibinfo{author}{Gurudu, S.~R.} \&
  \bibinfo{author}{Liang, J.}
\newblock \bibinfo{journal}{\bibinfo{title}{Automated polyp detection in
  colonoscopy videos using shape and context information}}.
\newblock {\emph{\JournalTitle{IEEE transactions on medical imaging}}}
  \textbf{\bibinfo{volume}{35}}, \bibinfo{pages}{630--644}
  (\bibinfo{year}{2015}).

\bibitem{jha2020kvasir}
\bibinfo{author}{Jha, D.} \emph{et~al.}
\newblock \bibinfo{title}{{Kvasir-SEG: A segmented polyp dataset}}.
\newblock In \emph{\bibinfo{booktitle}{International Conference on Multimedia
  Modeling}}, \bibinfo{pages}{451--462} (\bibinfo{year}{2020}).

\end{thebibliography}
\end{document}